\documentclass[acmsmall]{acmart}
\usepackage{graphicx}
\usepackage{makecell}
\usepackage{multirow}
\usepackage{tablefootnote}
\usepackage{amsmath,amssymb,amsfonts,bm}
\usepackage{latexsym}
\usepackage{subfigure}
\usepackage[misc]{ifsym} %
\usepackage{color}
\usepackage{algorithm}
\usepackage{algorithmic}

 \usepackage{multicol}

\AtBeginDocument{%
  \providecommand\BibTeX{{%
    \normalfont B\kern-0.5em{\scshape i\kern-0.25em b}\kern-0.8em\TeX}}}

\setcopyright{acmlicensed}
\acmJournal{TKDD}
\acmYear{2021} \acmVolume{1} \acmNumber{1} \acmArticle{1} \acmMonth{1} \acmPrice{15.00}\acmDOI{XXX/XXX}

\begin{document}

\title[DAPAMT]{Jointly Modeling Heterogeneous Student Behaviors and Interactions Among Multiple Prediction Tasks}

\author{Haobing Liu}
\email{liuhaobing@sjtu.edu.cn}

\author{Yanmin Zhu}
\email{yzhu@sjtu.edu.cn}

\author{Tianzi Zang}
\email{zangtianzi@sjtu.edu.cn}

\author{Yanan Xu}
\email{xuyanan2015@sjtu.edu.cn}

\author{Jiadi Yu}
\email{jiadiyu@sjtu.edu.cn}

\author{Feilong Tang}
\email{tang-fl@cs.sjtu.edu.cn}

\affiliation{%
  \institution{Department of Computer Science and Engineering, Shanghai Jiao Tong University}
  \city{Shanghai, China}
}

\renewcommand{\shortauthors}{Haobing L. et al.}

\begin{abstract}
Prediction tasks about students have practical significance for both student and college. Making multiple predictions about students is an important part of a smart campus. For instance, predicting whether a student will fail to graduate can alert the student affairs office to take predictive measures to help the student improve his/her academic performance. 
With the development of information technology in colleges, we can collect digital footprints which encode heterogeneous behaviors continuously.  
In this paper, we focus on modeling heterogeneous behaviors and making multiple predictions together, since some prediction tasks are related and learning the model for a specific task may have the data sparsity problem. 
To this end, we propose a variant of LSTM and a soft-attention mechanism. The proposed LSTM is able to learn the student profile-aware representation from heterogeneous behavior sequences. The proposed soft-attention mechanism can dynamically learn different importance degrees of different days for every student. In this way, heterogeneous behaviors can be well modeled. In order to model interactions among multiple prediction tasks, we propose a co-attention mechanism based unit. With the help of the stacked units, we can explicitly control the knowledge transfer among multiple tasks. 
We design three motivating behavior prediction tasks based on a real-world dataset collected from a college. Qualitative and quantitative experiments on the three prediction tasks have demonstrated the effectiveness of our model.
\end{abstract}

\ccsdesc[500]{Information systems~Data mining}
\ccsdesc[500]{Applied computing~Education}

\keywords{LSTM, Attention Mechanism, Heterogeneous Student Behaviors, Multi-Task Learning}

\maketitle

\section{Introduction}\label{in}
Recently, more and more people are concerned about the educational field. By utilizing data mining techniques in this field, there arise various significant prediction tasks for better understanding students and the settings which students learn in, such as academic performance prediction~\cite{yao2017predicting,xu2017progressive}, library circulation prediction~\cite{tian2011application,wang2012application}, graduation failure prediction~\cite{sukhbaatar2019artificial}. With the help of these tasks, educators could know grades of students, the library circulation, or whether students pass/fail for a specific given course ahead of time. Then educators could facilitate personalized education, do library strategic plan, or design in-time intervention.

Previous works about these prediction tasks mostly focus on the factors including values of historical observations (such as historical grades~\cite{xu2017progressive}, historical library circulation~\cite{tian2011application,wang2012application}) and student demographic information (i.e., student profiles)~\cite{shahiri2015review,nghe2007comparative}. These factors are relatively stable in the long run and are difficult to change via educational management. Besides, in online learning environments (e.g., massive open online courses), students' digital records collected by online learning platforms such as logs about video-watching behavior, time spent on specific questions, and test/quiz grades have been leveraged~\cite{brinton2015mooc,calvo2006predicting,romero2013predicting,lopez2012classification,minaei2003predicting}. The digital records can directly reflect students' efforts so they are important to prediction tasks. But these records are rarely digitized in traditional education.

Thanks to the development of information technology in colleges, there is a clear trend to augment physical facilities with sensing, computing, and communication capabilities~\cite{zhang2010extracting}. These facilities unobtrusively record students' digital footprints every day. The digital footprints of students encode heterogeneous behaviors that are helpful for prediction tasks. For example, academic efforts can be learned from entering the library records and entering the dormitory records. Academic efforts are key factors for predicting academic performance, the number of borrowed books, or the number of failed courses. In other words, students' hard study (i.e., entering the library frequently, early; going back to the dormitory late) can pay off and may result in borrowing many books from libraries. Once the digital footprints are available, they can be used to improve prediction performances.
\begin{figure}[t]
\centering
    \subfigure[]{
        \includegraphics[width= 6.5cm]{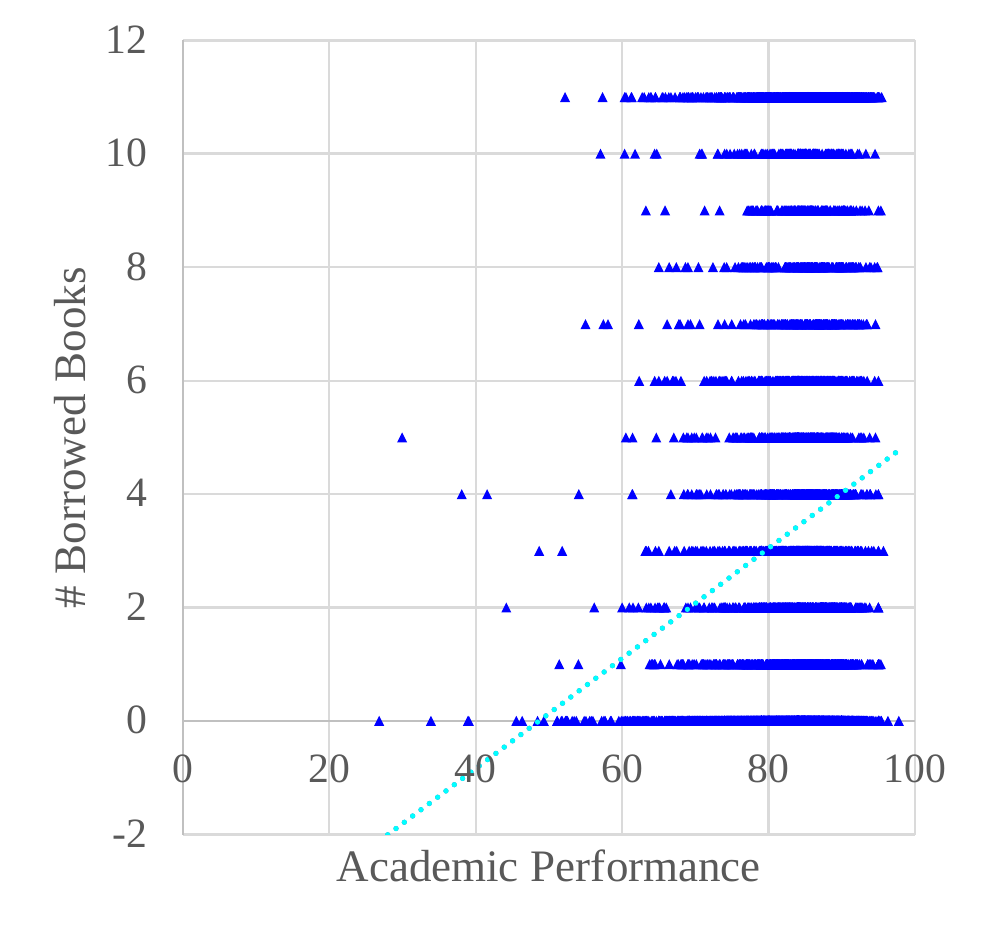}}
    \subfigure[]{
        \includegraphics[width= 6.5cm]{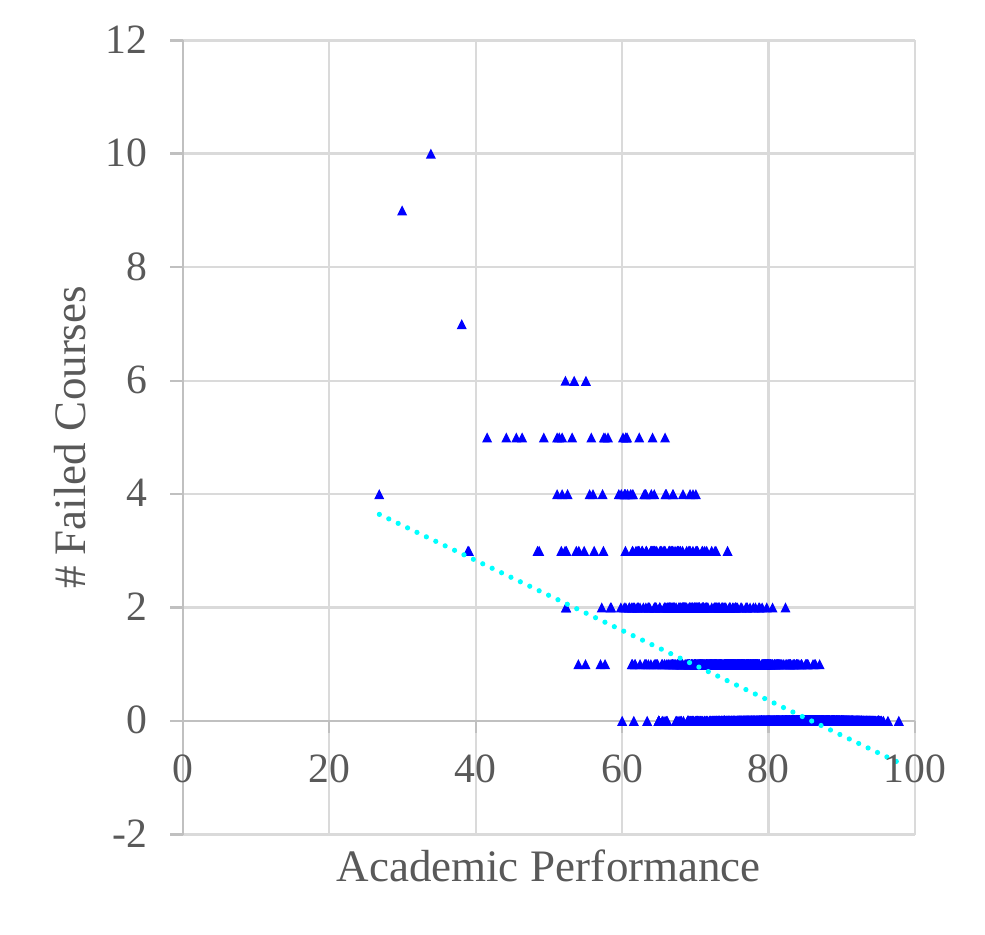}}
\caption{(a) shows the correlation between academic performance and \# borrowed books. (b) shows the correlation between academic performance and \# failed courses.}
\label{fig:correlation}
\end{figure}
Besides, some prediction tasks are closely related. Figure~\ref{fig:correlation} (a) shows the correlation between the academic performance and the number of borrowed books. The percentage of students who borrow more than $11$ books in one semester is low (around $13\%$) so we regard any number of borrowed books greater than $11$ as $11$. We can see that the academic performance and the number of borrowed books have a positive correlation. In Figure~\ref{fig:correlation} (b), the academic performance and the number of failed courses have a negative correlation. Hence, when predicting the number of books borrowed by student $s$, the value is more likely to be large once knowing that the academic performance of student $s$ is good and the number of failed courses is small.
In this paper, we collect digital footprints of $10k$ students spanning one academic year from campus smart card usage for entering the library and the dormitory. One footprint record mainly contains \emph{the student identity number} and \emph{the timestamp of the record}.

Based on the above thoughts, we focus on jointly modeling heterogeneous student behaviors generated from digital footprints and interactions among multiple prediction tasks. We take Predicting Academic Performance (PAP), Predicting the Number of Borrowed Books (PNBB), and Predicting the Number of Failed Courses (PNFC) as three motivating examples of prediction tasks. However, we mainly confront the following challenges.
First, the students' profile information has a large impact on their behaviors. Similar behaviors of different kinds of students may mean differently. For instance, there are two students: student $s$ and student $s^\prime$. Their records of entering the library are similar. Student $s$ is a freshman while student $s^\prime$ is a senior. Student $s^\prime$ would have less homework to do than student $s$. In return, it means student $s^\prime$ works harder than student $s$. Thus the academic performance of student $s^\prime$ should be better than student $s$. Profile information has not been well considered in previous studies. So the challenge is \emph{how to consider profile information while modeling daily behaviors}.
Second, behaviors of different days have different degrees of impact. For instance, the academic efforts of students on different days will change due to many reasons such as study habits. So the challenge is \emph{how to dynamically find out informative days according to different students}.
Third, a simple way to capture correlations among multiple tasks is leveraging a simple multi-task learning framework. In this way, we can only capture latent interactions and we can not explicitly control the knowledge transfer, resulting in a lack of interpretability. So the challenge is \emph{how to explicitly model interactions among multiple tasks}.

To address the first two challenges, we propose an attentional profile-aware multi-task model (APAMT) for modeling heterogeneous daily behaviors in our preliminary work~\cite{liu2020learning}. More specifically, for heterogeneous daily behaviors, each kind of behavior sequence is modeled by a variant of LSTM named Profile-aware LSTM. By adding profile information in the gates of LSTM, LSTM can consider profile information when modeling the behavior sequence, so as to improve the performances of prediction tasks. Besides, a novel soft-attention mechanism is designed over Profile-aware LSTM to dynamically learn the different importance degrees of different days for every student for improving the prediction performances. In APAMT model, we leverage a simple multi-task learning framework to implicitly model interactions among multiple tasks. Therefore, in this paper, we extend APAMT and propose a \underline{D}ual \underline{A}ttention \underline{P}rofile-\underline{A}ware \underline{M}ulti-\underline{T}ask model (DAPAMT) to explicitly model interactions among multiple tasks. Specifically, we design a unit called Multi-task Interaction Unit, the core of which is the co-attention mechanism. With the unit, we can explicitly control the process of knowledge transfer. Besides, we can stack more than one unit to model the high-level interactions. Moreover, we find that leveraging multi-task learning is necessary since there exists a data sparsity problem. For example, in our dataset, around $44\%$ of students do not borrow books in one semester. Thus, the input data of PNBB task can be sparse.
 
In summary, the main contributions of this paper are as follows:
\begin{itemize}
    \item We propose a Dual Attention Profile-Aware Multi-Task model (i.e., DAPAMT) to deal with multiple prediction tasks about students simultaneously. DAPAMT can jointly model heterogeneous student behaviors generated from digital footprints effectively and interactions among multiple prediction tasks explicitly.
    \item We design a variant of LSTM called Profile-aware LSTM to capture profile information when modeling the daily behavior sequence. We design a soft-attention mechanism to dynamically find out informative days. We design a co-attention mechanism based unit called Multi-task Interaction Unit to explicitly control the knowledge transfer and alleviate the data sparsity problem.
    \item We evaluate our proposed model on a large-scale real-world dataset. The experimental results demonstrate that our model outperforms competing baselines and every component of our model is well-designed, benefiting the prediction.
\end{itemize}

The rest of this paper is organized as follows. We first introduce the related work of our research in Section~\ref{rw}. Next, we introduce the problem statement in Section~\ref{pr}. Following that, we propose our model in Section~\ref{mo}. Then, Section~\ref{ex} presents qualitative and quantitative results of different methods. Section~\ref{ec} presents the ethical and social implications of this work. Finally, we conclude the paper in Section~\ref{co}.

\section{Related Work}\label{rw}
In this section, we discuss existing studies that are related to the three prediction tasks or the methods we used in this paper.
\subsection{RNNs for User Behavior Modeling}
Recurrent Neural Networks (RNNs) have been widely adopted to model sequence data and have achieved good performance in various domains such as NLP~\cite{sutskever2014sequence}. There are many well-known variants of RNN models, such as LSTM~\cite{hochreiter1997long}, GRU~\cite{chung2014empirical}, bidirectional LSTM~\cite{graves2005framewise}. Hidasi et al.~\cite{hidasi2016session} firstly introduced GRU to recommender systems. They leveraged GRU to model click behaviors of users. Zhu et al.~\cite{zhu2017next} proposed a variant of LSTM called Time-LSTM. They proposed that it is important to exploit the time information when modeling users’ behaviors. To achieve this goal, they equipped LSTM with time gates to model time intervals. 

\subsection{Academic Performance Predictions}
Academic performance prediction task is the most frequently studied among prediction tasks about students. In this paper, we choose Weighted Average Grade (WAG) which is on a 100-point scale to quantitatively describe the academic performance of a student in one semester. WAG can be seen as term GPA.

There are various factors that impact the academic performance in complex ways~\cite{khan2020student}. Student demographic information and historical academic performance are widely explored by researchers~\cite{shahiri2015review,khan2020student}. Some technologies such as online learning platforms are used in some courses nowadays. Students' digital records collected by online learning platforms such as logs about video-watching behavior, time spent on specific questions, and test/quiz grades have been leveraged~\cite{brinton2015mooc,calvo2006predicting,romero2013predicting,lopez2012classification,minaei2003predicting,khan2020student}. A few studies find that effective teaching motivates the student to perform better~\cite{khan2018data,khan2020student}. Besides, some studies explore other factors such as attitudes towards study~\cite{osmanbegovic2012data}. As for employing students' behaviors on campus to predict academic performance, Wang et al.~\cite{wang2015smartgpa} found correlations between students’ cumulative GPAs and automatic sensing behavioral data obtained from smartphones. However, the passive sensing behavioral data they used is only collected from a small number of students and the collecting way is not universal enough. Yao et al.~\cite{yao2017predicting} studied the effect of social influence on predicting academic performance based on students' multiple behaviors. The effect of students' behaviors is very indirect. Zhang et al.~\cite{zhang2018students} extracted statistics features and relevance features from students' multiple behaviors and used these features to predict academic performance.

Most methods used for academic performance prediction are based on traditional data mining techniques. Khan and Ghosh~\cite{khan2018data} regarded the student evaluation of teaching excellence and the number of student evaluations as input and used association rule mining to establish the relationship between teaching and academic performance. Feng et al.~\cite{feng2009addressing} did a stepwise linear regression to predict academic performance using the online measures as independent variables. Sweeney et al.~\cite{sweeney2015next} explored the factorization machine, a general-purpose matrix factorization algorithm based on (student, course) dyads. Xu et al.~\cite{xu2017progressive} leveraged an ensemble learning method to predict academic performance continuously in the program based on the evolving academic performances. Besides traditional methods, some researchers seek novel solutions with the help of deep learning. Wang and Liao~\cite{wang2011data} gathered information about gender, personality type, and anxiety level through questionnaires. Then they adopted a feed forward neural network to predict academic performance. LSTMs are also explored. Fei and Yeung~\cite{fei2015temporal} leveraged an LSTM to model student weekly activities and predict academic performance in Massive Open Online Courses (MOOCs).

These studies will ignore the influence of profile information when modeling student daily behaviors and treat all days equally.

\subsection{Book-Borrowing Predictions}


Book-borrowing predictions are meaningful and have been studied by many researchers. Cano et al.~\cite{cano2018case} analysed student book-borrowing behavior during the semester through clustering and association rule algorithms. They proposed that librarians could take better decisions (such as decide the quantity to offer per topic and human resources needed to satisfy the demand) for their users after knowing the analysis results. 

Besides, most existing studies on book-borrowing prediction focus on predicting library circulation. Tian~\cite{tian2011application} regarded the time series of library circulation flow as the chaotic time series and leveraged support vector regressions to predict library circulation of coming months. They mentioned that when the number of enrollments in a college increases sharply, the importance of book-borrowing predictions comes out conspicuously. Kumar and Raj~\cite{kumar2016improving} found that the number of borrowed books one month and twelve months earlier could estimate the number of borrowed books in a month with an autoregressive integrated moving average model. Wang et al.~\cite{wang2012application} leveraged a feed forward neural network to predict library circulation of next five days based on data of previous twenty days. These works proposed that they were of positive significance for library capacity planning, management of library books and staff, and library acquisition budget supporting. These works ignore the impact of individuals.

\subsection{Course-Failing Predictions}
Existing studies on course-failing prediction mainly focus on classifying students into two categories: either pass or fail for a given course. Existing studies do not distinguish this task from academic performance prediction task~\cite{yu2010feature}. If there is only one course, these two tasks are the same. Once we knew one student's mark, we know whether the student failed the course or not and vice-versa. If there is more than one course, these two tasks are different. For example, if student $s$ gets $80$ points in course $c_1$ ($3$ credits) and gets $50$ points in course $c_2$ ($3$ credits), the academic performance of student $s$ is $65$ which is above $60$ points. But, the number of failed courses of this student is $1$ (he/she fails course $c_2$). 

Actually, both the two tasks could be equally important. According to college rules of student management, either a low Weighted Average Grade or a certain number of failed courses lead to the academic probation even the academic dismissal. Thus, to effectively predict those students who may violate college academic requirements, both the academic performance prediction and the course-failing prediction are needed. Besides, some students may be more concerned with their Weighted Average Grades, because they may apply for graduate studies. Knowing their Weighted Average Grades previously would help them to study harder. Nevertheless, some students may be more concerned with whether they fail or not, because if they failed a course, they would not be qualified for student loans. In this case, knowing whether they fail or not would help them to act accordingly.

Methods used in course-failing prediction include k-nearest neighbour methods~\cite{tanner2010predicting}, ensemble methods~\cite{yu2010feature}, feed forward neural networks~\cite{sukhbaatar2019artificial} and so on. In detail, Tanner and Toivonen~\cite{tanner2010predicting} took student status features, student performance features, lesson requirement features, and customer demographic features as possible input of a k-nearest neighbor method to do course-failing prediction in an online course setting. Yu et al.~\cite{yu2010feature} did feature engineering and used a random forest algorithm (i.e., an ensemble method) to predict whether the student will be correct on the first attempt for a step based on interaction logs generated in intelligent tutoring systems. Sukhbaatar et al.~\cite{sukhbaatar2019artificial} employed a feed forward neural network on the set of prediction factors extracted from the online learning activities of students to identify students at risk of failing in a course.

\subsection{Multi-Task Learning}
Multi-task learning (MTL) was first analyzed by Caruana~\cite{caruana1997multitask} in detail. Multi-task learning could improve learning efficiency and prediction accuracy for each task when compared to training a separate model for each task. One important reason is that multi-task learning allows sharing of statistical strength and transferring of knowledge between related tasks. Multi-task learning has been used successfully in many fields, such as computer vision~\cite{dai2016instance,ranjan2017hyperface}, natural language processing~\cite{collobert2008unified}, urban computing~\cite{lu2020inferring}.

\section{Preliminaries}\label{pr}
In this section, we introduce the problem statement.

\noindent\textbf{Entering the Library Records.} When students enter the library, they need to swipe their campus cards. Thus records are generated. One entering the library record can be represented as $r_{Lib}=(s, t_{Lib})$, where $s$ denotes student $s$ and $t_{Lib}$ denotes the timestamp of the record.

\noindent\textbf{Entering the Dormitory Records.} Similar to entering the library records, one entering the dormitory record is represented as $r_{Dorm}=(s, t_{Dorm})$, where $s$ denotes student $s$ and $t_{Dorm}$ denotes the timestamp.

\noindent\textbf{Student Profile Records.} One record of this sub-dataset can be represented as $(s, \bm{D})$, where $\bm{D}$ is an attribute set about demographic information of student $s$. In this paper, according to the dataset, the attributes include the place of birth, nationality, gender, grade, school, and department. If other demographic information was available, it could also be added.

\noindent\textbf{Student Final Course Grade Records.} Given course $c_\theta$, one record can be represented as $r_{Grade}=(s, \mathcal{T}, c_\theta, credit(c_\theta), grade(s,c_\theta))$, where $\mathcal{T}$ is the index of the whole semesters that student $s$ involves in; $credit(c_\theta)$ is the credit of course $c_\theta$ and $grade(s,c_\theta)$ is the final grade student $s$ achieves in course $c_\theta$. $1\leq\mathcal{T}\leq T$.

\noindent\textbf{Problem Statement.} In this paper, we have three motivating tasks.

\textbf{PAP task:} Given digital footprints (i.e., $\{r_{Lib}\}$ and $\{r_{Dorm}\}$) generated in the first $X$ days of the semester $T\!+\!1$, profile information $\bm{D}$, and final course grade records $\{r_{Grade}\}$ generated in all previous $T$ semesters of student $s$, our goal is to predict the future academic performance (i.e., WGA) of student $s$ in semester $T\!+\!1$. $X\! <\! \#\ days\ in\ one\ semester$.

\textbf{PNBB task:} Given digital footprints (i.e., $\{r_{Lib}\}$ and $\{r_{Dorm}\}$) generated in the first $X$ days of the semester $T\!+\!1$, profile information $\bm{D}$, and the number of borrowed books in each previous semester (the number of all previous semesters is $T$) of student $s$, our goal is to predict the number of borrowed books of student $s$ in semester $T\!+\!1$. $X\! <\! \#\ days\ in\ one\ semester$.

\textbf{PNFC task:} Given digital footprints (i.e., $\{r_{Lib}\}$ and $\{r_{Dorm}\}$) generated in the first $X$ days of the semester $T\!+\!1$, profile information $\bm{D}$, and final course grade records $\{r_{Grade}\}$ generated in all previous $T$ semesters of student $s$, our goal is to predict the number of failed courses of student $s$ in semester $T\!+\!1$.  $X\! <\! \#\ days\ in\ one\ semester$.

\section{Proposed DAPAMT Model}\label{mo}
\begin{figure*}[t]
\centering
    \includegraphics[width= 13cm]{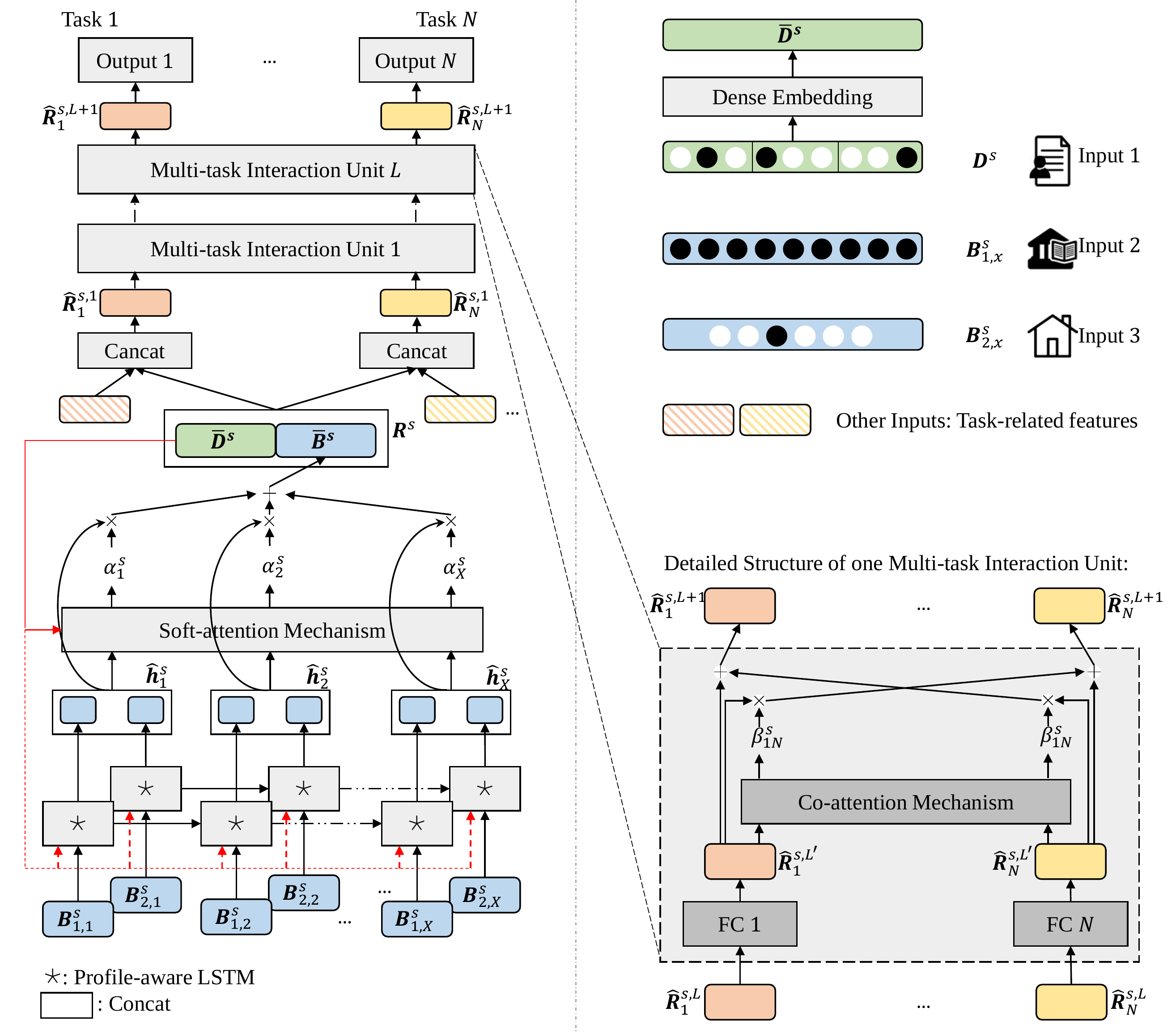}
    \caption{The main structure of DAPAMT.}
    \label{fig:structure}
\end{figure*}

The main structure of our proposed Dual Attention Profile-Aware Multi-Task model (DAPAMT) is illustrated in Figure~\ref{fig:structure}. The rest of this section is organized as follows. First, we introduce how to model the daily behavior sequence with Profile-aware LSTM and attention-based pooling. Then we introduce how to explicitly model the interaction among multiple prediction tasks with stacked Multi-task Interaction Units. Finally, we utilize task-specific output layers to get the final prediction results. Table~\ref{tab:not} summarizes main notations and their meanings used throughout this paper.

\begin{table}[t]
\caption{Main notations.}
\small
\label{tab:not}
\centering
\begin{tabular}{l|l}
\hline
Notation                                                                                                     & Description                                                                                                                                    \\ \hline
$\bm{D}^s$                                                                                                   & Profile information of student $s$.                                                                                                            \\
$\bm{\bar{D}}^s$                                                                                             & Dense representation of the profile information.                                                                                               \\
$\bm{B}_{m,x}$                                                                                               & Feature vector of the $m$-th kind of behavior at the $x$-th day.                                                                               \\
$\bm{i}$, $\bm{f}$ and $\bm{o}$                                                                              & Input, forget and output gates.                                                                                                                \\
$\bm{h}_{m,x}^s$                                                                                             & \begin{tabular}[c]{@{}l@{}}Hidden state of Profile-aware LSTM that models the $m$-th kind of behavior \\ at the $x$-th timestamp.\end{tabular} \\
$\bm{\hat{h}}_x^s$                                                                                           & Hidden representation of all heterogeneous behaviors of the $x$-th day.                                                                        \\
$\alpha$ terms                                                                                               & Soft-attention weights.                                                                                                                        \\
$\bm{\bar{B}}^s$                                                                                             & Advanced student behavior representation.                                                                                                      \\
$\bm{R}^s$                                                                                                   & Student representation.                                                                                                                        \\
$y_{1,\mathcal{T}}^s$, $y_{2,\mathcal{T}}^s$ and $y_{3,\mathcal{T}}^s$                                                   & WAG, \# borrowed books and \# failed courses in semester $\mathcal{T}$.                                                                        \\
$\bm{\hat{R}}_{1}^{s,l}$, $\bm{\hat{R}}_{2}^{s,l}$ and $\bm{\hat{R}}_{3}^{s,l}$                      & Inputs (task-specific feature vectors) of the $l$-th Multi-task Interaction Unit.                                                                  \\
$\bm{\hat{R}}_{1}^{s,l^\prime}$, $\bm{\hat{R}}_{2}^{s,l^\prime}$ and $\bm{\hat{R}}_{3}^{s,l^\prime}$ & More task-specific feature vectors.                                                                                                            \\
$\beta$ terms                                                                                                & Co-attention weights.                                                                                                                          \\ \hline
\end{tabular}
\end{table}

\subsection{Inputs and Dense Embedding Layer}
Student $s$' profile information $\bm{D}^s$ is represented in the form of one high-dimensional vector including many one-hot encoded vectors. To reduce the dimension and get a better representation, we use a dense embedding layer. The transformation is formalized as:
\begin{equation}
    \bm{\bar{D}}^s=\bm{W}_{D}\bm{D}^s,
    \label{de}
\end{equation}
where $\bm{W}_{D}$ is the mapping matrix.

Thinking that the combination of entering the library and going back to the dormitory behaviors can reveal students' academic efforts, we extract the daily behavior sequence of entering the library and the daily behavior sequence of going back to the dormitory from $\{r_{Lib}\}$ and $\{r_{Dorm}\}$ respectively.

More specifically, we divide one day into $24$ time slots by hour (i.e., $[00\!:\!00,01\!:\!00),[01\!:\!00,02\!:\!00),...,[23\!:\!00,24\!:\!00)$). According to $\{r_{Lib}\}$, almost 100\% of entering the library records are generated in 07:00-23:00. So we use $16$ elements to record entering the library frequency in 07:00-23:00 for each day. In this way, we get the daily behavior sequence of entering the library: $\bm{B}_{1,1},\bm{B}_{1,2},...,\bm{B}_{1,x},...,\bm{B}_{1,X}$ ($x$ is the index. $X\!=\!63$ since we make predictions after the first half of the semester (63 days).).

One going back to the dormitory record is defined as the last record of entering the dormitory of the day. Based on $\{r_{Dorm}\}$, around 84\% of going back to the dormitory records are generated in 18:00-24:00. So we use a vector with a length of $6$ to record the situation of going back to the dormitory in 18:00-24:00 for each day. In this way, we get the behavior sequence of going back to the dormitory: $\bm{B}_{2,1},\bm{B}_{2,2},...,\bm{B}_{2,x},...,\bm{B}_{2,X}$.

\subsection{Profile-aware LSTM}
Similar behaviors of different kinds of students may mean differently. We propose a variant of LSTM called Profile-aware LSTM. We treat student profile information as a strong signal in the gates of Profile-aware LSTM (as Equation~\eqref{input}, \eqref{forget} and \eqref{output} show). That is to say, what to extract, what to remember, and what to forward are affected by student profile information. The behavior sequence is the input of Profile-aware LSTM (as Equation~\eqref{onlyinput} shows).

Profile-aware LSTM model is formulated as follows and the detailed structure is shown in Figure~\ref{fig:plstm}:

\begin{figure}[t]
\centering
    \includegraphics[width= 8cm]{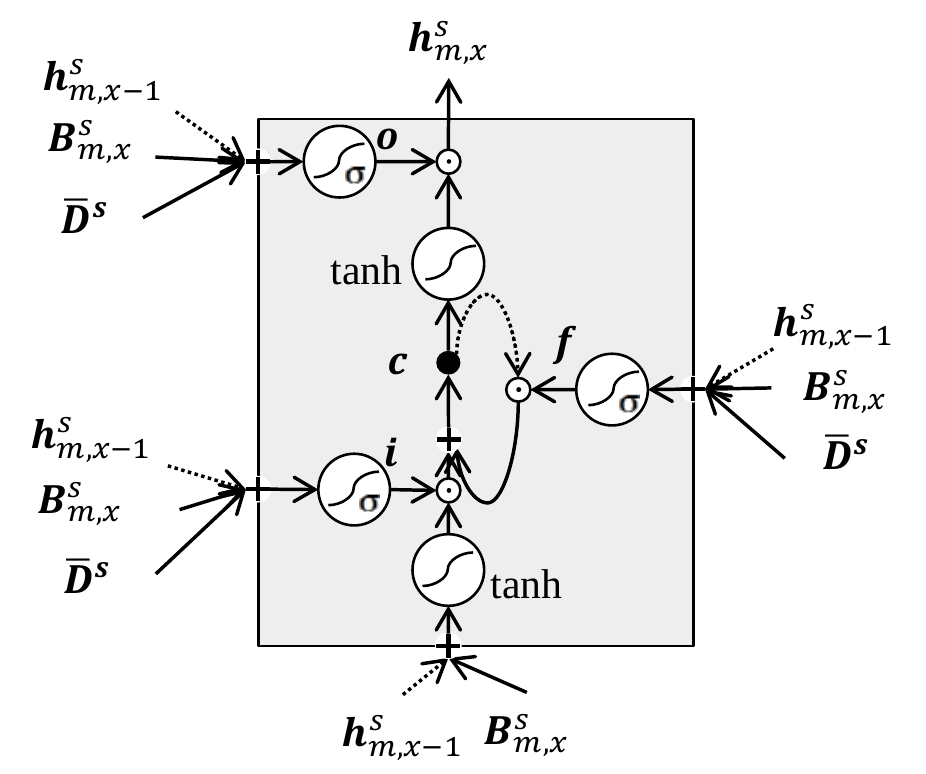}
    \caption{Detailed structure of Profile-aware LSTM.}
    \label{fig:plstm}
\end{figure}

\begin{align}
    &\bm{i}_{m,x}=\sigma(\bm{W}_{iB}\bm{B}_{m,x}^s+\bm{W}_{ih}\bm{h}_{m,x-1}^s+\bm{W}_{iD}\bm{\bar{D}}^s+\bm{b}_i), \label{input} \\
    &\bm{f}_{m,x}=\sigma(\bm{W}_{fB}\bm{B}_{m,x}^s+\bm{W}_{fh}\bm{h}_{m,x-1}^s+\bm{W}_{fD}\bm{\bar{D}}^s+\bm{b}_f), \label{forget} \\
    &\bm{c}_{m,x}=\bm{f}_{m,x}\odot\bm{c}_{m,x-1}+\bm{i}_{m,x}\odot \tanh(\bm{W}_{cB}\bm{B}_{m,x}^s+\bm{W}_{ch}\bm{h}_{m,x-1}^s+\bm{b}_c), \label{onlyinput} \\
    &\bm{o}_{m,x}=\sigma(\bm{W}_{oB}\bm{B}_{m,x}^s+\bm{W}_{oh}\bm{h}_{m,x-1}^s+\bm{W}_{oD}\bm{\bar{D}}^s+\bm{b}_o), \label{output} \\
    &\bm{h}_{m,x}^s=\bm{o}_{m,x}\odot \tanh(\bm{c}_{m,x}),
\end{align}
where $\bm{B}_{m,x}^s$ and $\bm{h}_{m,x}^s$ are one input element and the corresponding output of Profile-aware LSTM unit, i.e., hidden state at time $x$, respectively. $\bm{\bar{D}}^s$ is calculated with Equation~\eqref{de}. $m=1,2$. $1\!\leq\! x\!\leq\! X\!=\!63$. $\bm{W}$ terms denote weight matrices and $\bm{b}$ terms are bias vectors. $\sigma$ is the element-wise sigmoid function and $\odot$ is the element-wise product.

We use two Profile-aware LSTMs to model two kinds of behaviors respectively.

\subsection{Attention-based Pooling Layer}
The hidden representation of all heterogeneous behaviors of the $x$-th day can be formalized as:
\begin{equation}
    \bm{\hat{h}}_x^s=\bm{h}_{1,x}^s\oplus\bm{h}_{2,x}^s,
    \label{concat1}
\end{equation}
where $\oplus$ is the concatenation operation.

Behaviors of different days to the task will have different degrees of impact. Inspired by the success of the attention mechanism in machine translation~\cite{bahdanau2014neural}, we apply a novel soft-attention mechanism over Profile-aware LSTM to draw information from the sequence by different weights. In detail, we consider each vector $\bm{\hat{h}}_x^s$ as heterogeneous behaviors representation of the $x$-th day, and represent the sequence by a weighted sum of the vector representation of all days. The attention weight makes it possible to perform proper credit assignments to days according to their importance to the student. Mathematically, we compute soft-attention weight $\alpha_x^s$ with the following equations:

\begin{align}
    a_x^s &=\bm{W}_{a0}\tanh(\bm{W}_{a1}\bm{\hat{h}}_x^s+\bm{W}_{a2}\bm{\bar{D}}^s+\bm{b}_{a}),\\
    \alpha_x^s &=\frac{\exp(a_x^s)}{\sum_{x^\prime=1}^X \exp(a_{x^\prime}^s)},
\end{align}
where $\bm{\bar{D}}^s$ is computed with Equation~\eqref{de}, $\bm{\hat{h}}_x^s$ is computed with Equation~\eqref{concat1}, $\bm{W}$ terms denote weight matrices, and $\bm{b}_a$ is the bias vector. Similar to Profile-aware LSTM, student profile information also contributes to the attention weights.

The advanced student behavior representation is generated using the following equation:
\begin{equation}
    \bm{\bar{B}}^s=\sum_{x=1}^X \alpha_x^s \bm{\hat{h}}_x^s.
\end{equation}

The student representation $\bm{R}^s$ is the concatenation of $\bm{\bar{D}}^s$ and $\bm{\bar{B}}^s$:
\begin{equation}
    \bm{R}^s=\bm{\bar{D}}^s\oplus\bm{\bar{B}}^s.
\end{equation}

\subsection{Stacked Multi-task Interaction Units}\label{mt}
\begin{figure*}[t]
\centering
    \includegraphics[width= 12.5cm]{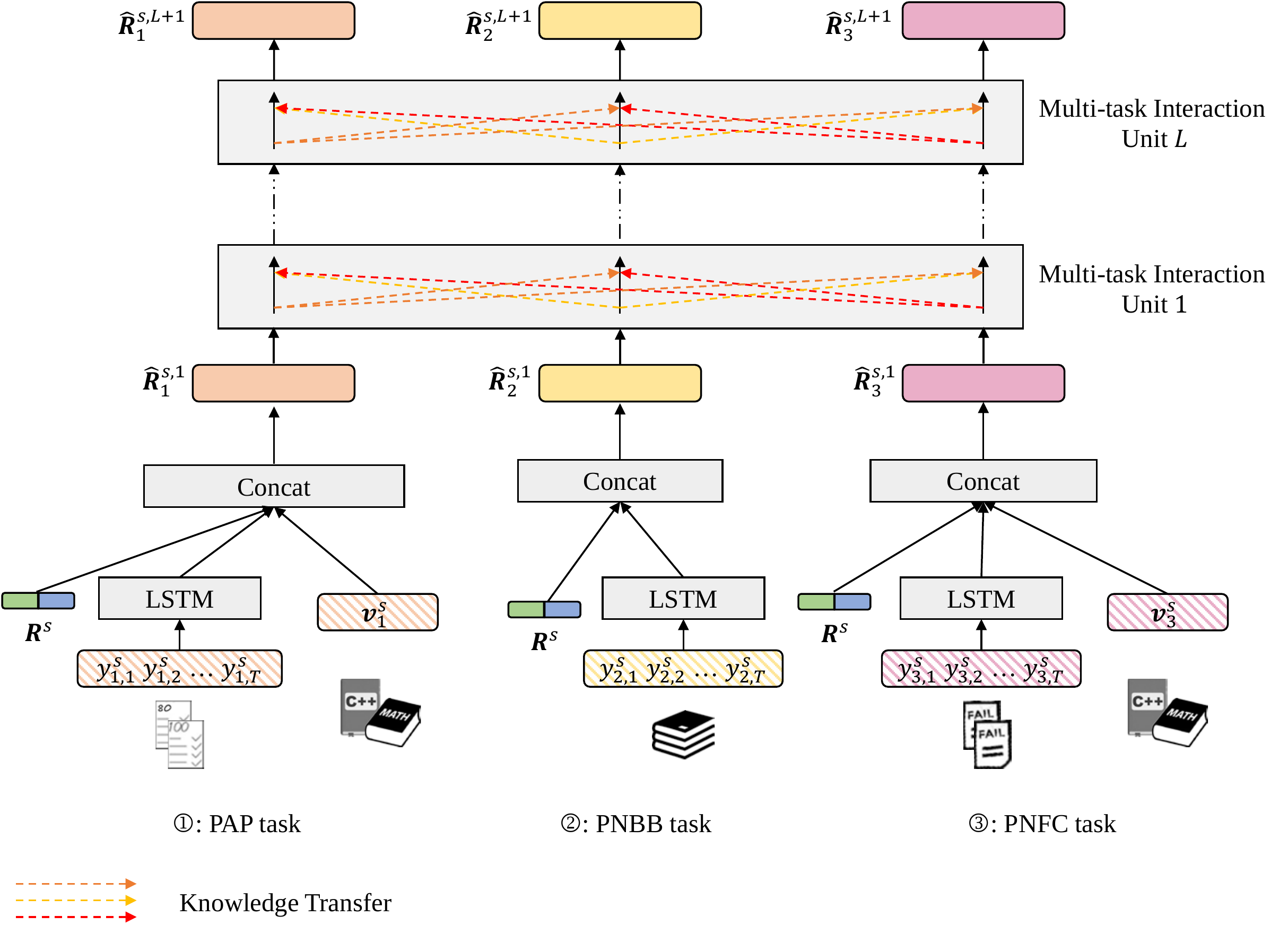}
    \caption{Detailed structure of stacked Multi-task Interaction Units with the three motivating tasks.}
    \label{fig:tasks}
\end{figure*}

After obtaining the student representation, we concatenate it with task-related features. In detail, we have three tasks: PAP task, PNBB task, and PNFC task. For PAP task, the task-related information includes the historical WAG information and the involved course information; for PNBB task, the task-related information includes the historical number of borrowed books information; for PNFC task, the task-related information includes the historical number of failed courses and the involved course information as Figure~\ref{fig:tasks} shows. Then we feed the concatenated features to stacked Multi-task Interaction Units. In what follows, we introduce how to handle the various kinds of task-related information and introduce the detailed structure of the Multi-task Interaction Unit.

\noindent\textbf{PAP Task.} As mentioned in Section~\ref{rw}, we use WAG to quantitatively describe the academic performance. Given $\{r_{Grade}\}$, WAG is calculated with the following equation:
\begin{equation}
    y_{1,\mathcal{T}}^s=\sum_{\theta=1}^{\Theta} \frac{credit(c_\theta)grade(s,c_\theta)}{\sum_{\theta=1}^{\Theta}credit(c_\theta)},
\end{equation}
where $\Theta$ is the number of courses chosen by student $s$ in semester $\mathcal{T}$. In this way, we get the historical WAG sequence of student $s$: $y_{1,1}^s,y_{1,2}^s,...,y_{1,\mathcal{T}}^s,...,y_{1,T}^s$. The historical WAG sequence reveals students' trends of academic performance. Note that the length of the historical WAG sequence may vary from student to student, so we adopt a dynamic LSTM to model the sequence:
\begin{equation}
    \bm{\hbar}_{1,\mathcal{T}}^s=\mathrm{LSTM}(y_{1,\mathcal{T}}^s,\bm{\hbar}_{1,\mathcal{T}-1}^s),\quad 1\leq\mathcal{T}\leq T,
\end{equation}
where $y_{1,\mathcal{T}}^s$ and $\bm{\hbar}_{1,\mathcal{T}}^s$ are one input and the corresponding hidden state at time $\mathcal{T}$. We regard the last hidden state $\bm{\hbar}_{1,T}^s$ as the trend representation.

\begin{figure*}[t]
\centering
    \includegraphics[width= 11.5cm]{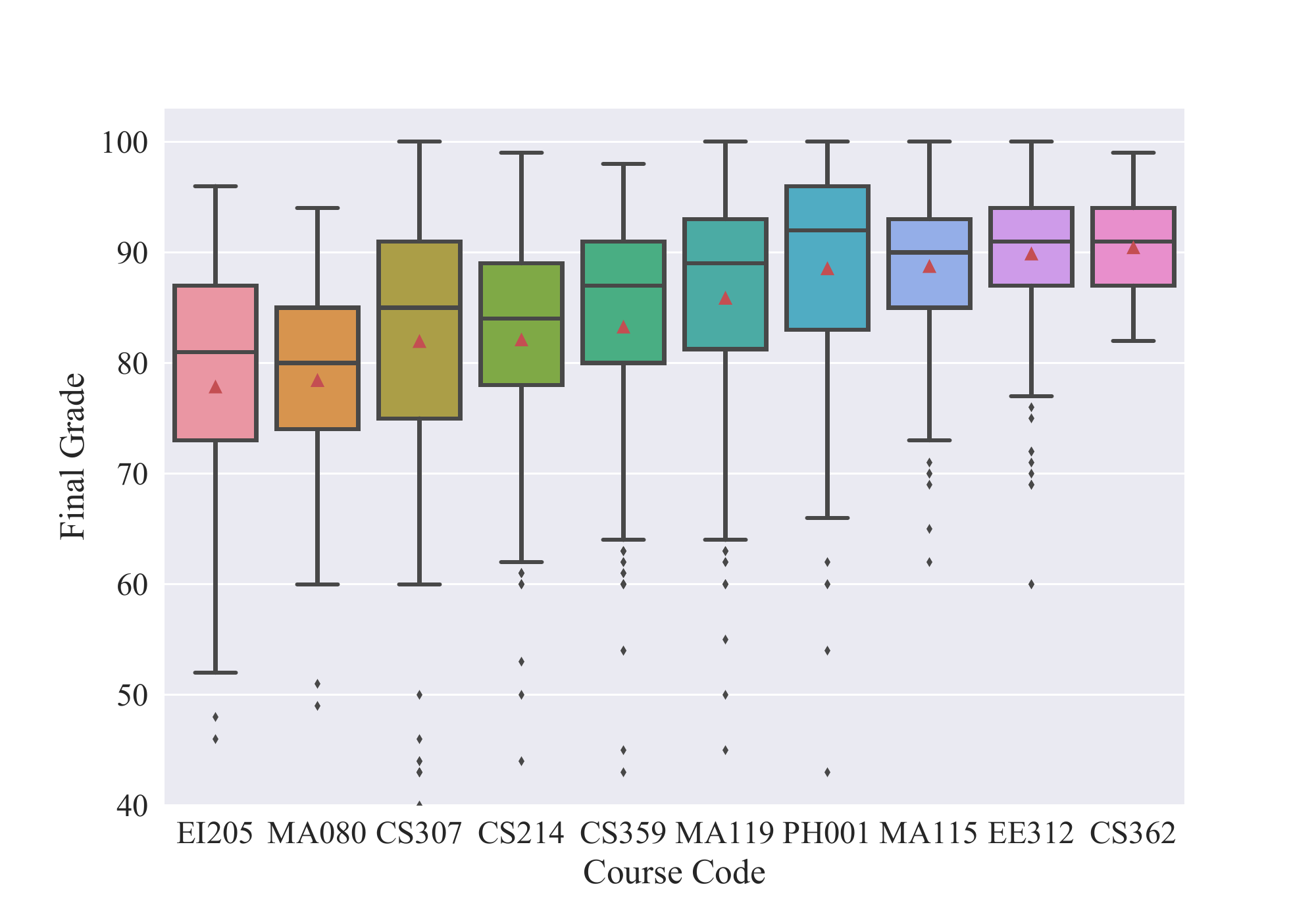}
    \caption{Grade distributions of some different courses set in Department of Computer Science and Engineering.}
    \label{fig:courseGrade}
\end{figure*}

We find that students get higher grades easily in some courses such as CS362 as Figure~\ref{fig:courseGrade} shows. This means different courses have different levels of difficulty. So for each course, we extract descriptive statistics (i.e., minimum, maximum, median, first quartile, third quartile, mean, standard deviation) as features. These features can describe the properties of distribution from multiple aspects. The feature vector of course $c_\theta$ is represented as $\bm{e}_{1,\theta}$. Next, we aggregate feature vectors of all courses by leveraging course credit information:
\begin{equation}
    \bm{v}_1^s=\sum_{\theta=1}^\Theta \frac{credit(c_\theta)\bm{e}_{1,\theta}}{\sum_{\theta=1}^{\Theta}credit(c_\theta)}.
\end{equation}

Then we concatenate $\bm{R}^s$, $\bm{\hbar}_{1,T}^s$ with $\bm{v}_1^s$:
\begin{equation}
    \bm{\hat{R}}_{1}^s=\bm{R}^s\oplus \bm{\hbar}_{1,T}^s\oplus \bm{v}_1^s.
    \label{rpap}
\end{equation}

\noindent\textbf{PNBB Task.} Similarly, the historical number of borrowed books sequence reveals students' trends of the number of borrowed books. We utilize another dynamic LSTM to model the sequence:
\begin{equation}
    \bm{\hbar}_{2,\mathcal{T}}^s=\mathrm{LSTM}(y_{2,\mathcal{T}}^s,\bm{\hbar}_{2,\mathcal{T}-1}^s),\quad 1\leq\mathcal{T}\leq T,
\end{equation}
where $y_{2,\mathcal{T}}^s$ and $\bm{\hbar}_{2,\mathcal{T}}^s$ are one input and the corresponding hidden state at time $\mathcal{T}$. $\bm{\hbar}_{2,T}^s$ is the trend representation.

Then we concatenate $\bm{R}^s$ with $\bm{\hbar}_{2,T}^s$:
\begin{equation}
    \bm{\hat{R}}_{2}^s=\bm{R}^s\oplus \bm{\hbar}_{2,T}^s.
    \label{rpnbb}
\end{equation}

\noindent\textbf{PNFC Task.} A LSTM is leveraged to model the historical number of failed courses sequence:
 \begin{equation}
    \bm{\hbar}_{3,\mathcal{T}}^s=\mathrm{LSTM}(y_{3,\mathcal{T}}^s,\bm{\hbar}_{3,\mathcal{T}-1}^s),\quad 1\leq\mathcal{T}\leq T,
\end{equation}
where $y_{3,\mathcal{T}}^s$ and $\bm{\hbar}_{3,\mathcal{T}}^s$ are one input and the corresponding hidden state at time $\mathcal{T}$.

We also extract some features such as course failure rate and descriptive statistics. The feature vector of course $c_\theta$ is represented as $\bm{e}_{3,\theta}$. We merger information of every course by:
\begin{equation}
\bm{v}_3^s=\sum_{\theta=1}^\Theta \frac{\bm{e}_{3,\theta}}{\Theta}.
\end{equation}

Then we concatenate $\bm{R}^s$, $\bm{\hbar}_{3,T}^s$ with $\bm{v}_3^s$:
\begin{equation}
    \bm{\hat{R}}_{3}^s=\bm{R}^s\oplus \bm{\hbar}_{3,T}^s\oplus \bm{v}_3^s.
    \label{rpnfc}
\end{equation}

\noindent\textbf{Multi-task Interaction Unit.} We stack $L$ Multi-task Interaction Units. Formally, for the $l$-th unit, the inputs are $\bm{\hat{R}}_{1}^{s,l}$, $\bm{\hat{R}}_{2}^{s,l}$, and $\bm{\hat{R}}_{3}^{s,l}$. The outputs are $\bm{\hat{R}}_{1}^{s,l+1}$, $\bm{\hat{R}}_{2}^{s,l+1}$, and $\bm{\hat{R}}_{3}^{s,l+1}$. $1\leq l \leq L$.

Firstly, we utilize Fully Connected (FC) layers to make the inputs more task-specific and to let the inputs have the same dimensions, which can be written as:
\begin{align}
    \bm{\hat{R}}_{1}^{s,l^\prime}&=\mathrm{PReLU}(\bm{W}_{1}^l \bm{\hat{R}}_{1}^{s,l}+\bm{b}_{1}^l), \\
    \bm{\hat{R}}_{2}^{s,l^\prime}&=\mathrm{PReLU}(\bm{W}_{2}^l \bm{\hat{R}}_{2}^{s,l}+\bm{b}_{2}^l), \\
    \bm{\hat{R}}_{3}^{s,l^\prime}&=\mathrm{PReLU}(\bm{W}_{3}^l \bm{\hat{R}}_{3}^{s,l}+\bm{b}_{3}^l),
\end{align}
where $\bm{W}$ terms and $\bm{b}$ terms are learnable parameters, and $\mathrm{PReLU}$~\cite{he2015delving} is the nonlinear activation function. $\bm{\hat{R}}_{1}^{s,1}$, $\bm{\hat{R}}_{2}^{s,1}$, and $\bm{\hat{R}}_{3}^{s,1}$ are calculated with Equation~\eqref{rpap}, \eqref{rpnbb}, and \eqref{rpnfc}, respectively.

Then we utilize the co-attention mechanism to explicitly control the knowledge transfer. The co-attention mechanism was first proposed to jointly reasons about visual attention and question attention~\cite{lu2016hierarchical}. The co-attention mechanism demonstrates the ability of capturing the complementary important information. With the co-attention mechanism, we can get the updated representation explicitly incorporating other tasks' knowledge. Mathematically, we compute co-attention weights with the following equations:
\begin{align}
    \beta_{12}^s=&\sigma((\bm{\hat{R}}_{1}^{s,l^\prime})^\top \bm{\hat{R}}_{2}^{s,l^\prime}), \\
    \beta_{13}^s=&\sigma((\bm{\hat{R}}_{1}^{s,l^\prime})^\top \bm{\hat{R}}_{3}^{s,l^\prime}), \\
    \beta_{23}^s=&\sigma((\bm{\hat{R}}_{2}^{s,l^\prime})^\top \bm{\hat{R}}_{3}^{s,l^\prime}),
\end{align}
where $\sigma$ is the sigmoid function. That is to say, we leverage dot product and sigmoid function to calculate the co-attention weights.

The updated representations are generated using the following equations:
\begin{align}
    \bm{\hat{R}}_{1}^{s,l+1}=& \bm{\hat{R}}_{1}^{s,l^\prime}+\beta_{12}^s\bm{\hat{R}}_{2}^{s,l^\prime}+ \beta_{13}^s\bm{\hat{R}}_{3}^{s,l^\prime}, \\
    \bm{\hat{R}}_{2}^{s,l+1}=& \bm{\hat{R}}_{2}^{s,l^\prime}+\beta_{12}^s\bm{\hat{R}}_{1}^{s,l^\prime}+ \beta_{23}^s\bm{\hat{R}}_{3}^{s,l^\prime}, \\
    \bm{\hat{R}}_{3}^{s,l+1}=& \bm{\hat{R}}_{3}^{s,l^\prime}+\beta_{13}^s\bm{\hat{R}}_{1}^{s,l^\prime}+ \beta_{23}^s\bm{\hat{R}}_{2}^{s,l^\prime}.
\end{align}

\subsection{Output Layers}
After $L$ Multi-task Interaction Units, we can get the outputs of the last unit: $\bm{\hat{R}}_{1}^{s,L+1}$, $\bm{\hat{R}}_{2}^{s,L+1}$, and $\bm{\hat{R}}_{3}^{s,L+1}$. Then we adopt separate output layers to get the prediction results, which can be denoted as follows:
\begin{align}
    \tilde{y}_{1,T+1}^s=&\tanh(\bm{W}_{O1}\bm{\hat{R}}_{1}^{s,L+1}+b_{O1}), \\
    \tilde{y}_{2,T+1}^s=&\tanh(\bm{W}_{O2}\bm{\hat{R}}_{2}^{s,L+1}+b_{O2}), \\
    \tilde{y}_{3,T+1}^s=&\tanh(\bm{W}_{O3}\bm{\hat{R}}_{3}^{s,L+1}+b_{O3}),
\end{align}
where $\bm{W}$ terms and $b$ terms are parameters.

\begin{algorithm}[t]  
	\caption{DAPAMT Training Algorithm}
	\label{trainalg}
	\small
	\begin{multicols}{2}
	\begin{algorithmic}[1] 
    \REQUIRE ~~\\ 
        For each student: given $M$ daily behavior sequences of length $X$ generated in semester $T+1$ $\{\bm{B}_{1,1},...,\bm{B}_{1,X};...;\bm{B}_{M,1},...,\bm{B}_{M,X}\}$; student profile information $\bm{D}$; $N$ historical observation sequences of length $T$ $\{y_{1,1},...,y_{1,T};...;y_{N,1},...,y_{N,T}\}$; additional task-related feature vectors $\{\bm{v}_{1},...,\bm{v}_{N}\}$ (some tasks may not have corresponding additional task-related feature vectors). The number of Multi-task Interaction Units $L$. $N$ balance weights $\{\lambda_1,...,\lambda_N\}$\\
        
    \ENSURE ~~\\ 
        Learned DAPAMT model
    \STATE \emph{//Forward propagation}
    \STATE Initialization;
    \STATE \emph{//Dense Embedding Layer}
    \STATE $\bm{\bar{D}}=\bm{W}_{D}\bm{D}$;
    \STATE \emph{//Profile-aware LSTMs}
    \FOR{each $m \in [1,M]$}
        \STATE $\bm{h}_{m,x}=$ Profile-aware\_LSTM($\bm{B}_{m,x},\bm{\bar{D}}$), $1\leq x\leq X$;\
    \ENDFOR
    \STATE \emph{//Attention-based Pooling Layer}
    \FOR{each $x \in [1,X]$}
        \STATE $\bm{\hat{h}}_x=\bm{h}_{1,x}\oplus\cdots\bm{h}_{m,x}\cdots\oplus\bm{h}_{M,x}$;\
    \ENDFOR
    \FOR{each $x \in [1,X]$}
        \STATE $\alpha_x=$ Soft-attention\_Mechanism($\bm{\hat{h}}_x,\bm{\bar{D}}$);\
    \ENDFOR
    \STATE $\bm{\bar{B}}\leftarrow\bm{0}$
    \FOR{each $x \in [1,X]$}
        \STATE $\bm{\bar{B}}\leftarrow\bm{\bar{B}}+\alpha_x\bm{\hat{h}}_x$;\
    \ENDFOR
    \STATE $\bm{R}=\bm{\bar{D}}\oplus\bm{\bar{B}}$
    \STATE \emph{//Stacked Multi-task Interaction Units}
    \FOR{each $n \in [1,N]$}
        \STATE $\bm{\hbar}_{n,\mathcal{T}}=$ LSTM($y_{n,\mathcal{T}}$), $1\leq\mathcal{T}\leq T$;\
    \ENDFOR
    \FOR{each $n \in [1,N]$}
        \STATE $\bm{\hat{R}}_{n}^{1}\leftarrow\bm{\hat{R}}_{n}=\bm{R}\oplus\bm{\hbar}_{n,T}\oplus\bm{v}_n$ (if $\bm{v}_n$ exists);\
    \ENDFOR 
    \FOR{each $l \in [1,L]$}
        \FOR{each $n \in [1,N]$}
            \STATE $\bm{\hat{R}}_{n}^{l^\prime}=\mathrm{PReLU}(\bm{W}_{n}^l \bm{\hat{R}}_{n}^{l}+\bm{b}_{n}^l)$;\
        \ENDFOR
        \FOR{each $i \in [1,N-1]$}
            \FOR{each $j \in [i+1,N]$}
                \STATE $\beta_{ij}=$ Co-attention\_Mechanism($\bm{\hat{R}}_{i}^{l^\prime},\bm{\hat{R}}_{j}^{l^\prime}$);\
            \ENDFOR
        \ENDFOR
        \FOR{each $n \in [1,N]$}
            \STATE $\bm{\hat{R}}_{n}^{l+1}=\bm{\hat{R}}_{n}^{l\prime}+\sum_{(1\leq i\leq N) \wedge (i\ne n)}\beta_{in}\bm{\hat{R}}_{i}^{l\prime}$;\
        \ENDFOR
    \ENDFOR
    \STATE \emph{//Output Layer}
    \FOR{each $n \in [1,N]$}
        \STATE $\tilde{y}_{n,T+1}=\tanh(\bm{W}_{On}\bm{\hat{R}}_{n}^{L+1}+b_{On})$;\
    \ENDFOR        
    \STATE \emph{//Back propagation}
    \STATE $\mathcal{L}=\sum_{n=1}^N \lambda_n$ MSE($y_{n,T+1},\tilde{y}_{n,T+1}$);
	\end{algorithmic}
	\end{multicols}
\end{algorithm}

\subsection{Optimization}
We use the mean squared error (MSE) as the loss function for training the three tasks:
\begin{align}
   \mathcal{L}_1(\Phi_1)=&\frac{1}{U}\sum_{i=1}^U(y^i_{1,T+1}-\tilde{y}^i_{1,T+1})^2, \\
   \mathcal{L}_2(\Phi_{2})=&\frac{1}{U}\sum_{i=1}^U(y^i_{2,T+1}-\tilde{y}^i_{2,T+1})^2, \\
   \mathcal{L}_3(\Phi_{3})=&\frac{1}{U}\sum_{i=1}^U(y^i_{3,T+1}-\tilde{y}^i_{3,T+1})^2, \\
\end{align}
where $i$ denotes one training sample, $U$ is the number of training samples, and $\Phi$ terms are all trainable parameters. $y^i_{1,T+1}$, $y^i_{2,T+1}$, and $y^i_{3,T+1}$ denote the labels of the $i$-th sample. 

The total loss is computed as the sum of the three individual losses:
\begin{equation}
\mathcal{L}_{Total}=\lambda_1\mathcal{L}_1+\lambda_2\mathcal{L}_2+\lambda_3\mathcal{L}_3,
\end{equation}
where $\lambda$ terms are balance weights. Here, balance weights of 1 are implicitly used among the three tasks.

The training process is outlined in Algorithm~\ref{trainalg}. We adopt the adaptive moment estimation (Adam)~\cite{kingma2014adam} as the optimizer. 

In order to improve the generalization capability of our models, we adopt dropout~\cite{srivastava2014dropout} to prevent the potential overfitting problem.

\section{Experiments}\label{ex}
In this section, we describe the detailed experimental settings and discuss the results.
\subsection{Dataset}
The data were collected in a college with an enrollment of $10k$ undergraduate students. Dataset statistics are shown in Table~\ref{tab:stat}.


\subsection{Comparison Baselines}
To demonstrate the effectiveness of our proposed model, we compared our proposed model, i.e., \textsf{DAPAMT}, with various baselines.

\begin{itemize}
    \item \textsf{HA}: We give the prediction result by the average value of historical observations.
    \item \textsf{LSTM}~\cite{fei2015temporal}: LSTM is widely used to model sequence data. We leverage LSTMs to model historical observation sequences.
    \item \textsf{BRR}~\cite{wang2015smartgpa}: Bayesian Ridge Regression (BRR) is a generalized linear model which has $L_2$ regularization. 
    \item \textsf{SVR}~\cite{tian2011application}: Support Vector Regression (SVR) is a variant of support vector machine (SVM) for supporting regression tasks. SVR is a minimum-margin regression which could model the non-linear relation between features.
    \item \textsf{RF}~\cite{chen2018early}: Random Forest (RF) is an ensemble method with decision trees as base learners. It is based on the ``bagging'' idea.
    \item \textsf{GBDT}: Gradient Boosting Decision Tree (GBDT) is another kind of ensemble method using decision trees as base learners. GBDT is a generalization of boosting to arbitrary differentiable loss functions.
    \item \textsf{MLP}~\cite{wang2011data,sukhbaatar2019artificial}: Multi-layer Perceptron (MLP) consists of multiple layers of nodes. Each layer is fully connected to the next layer in the network.
    \item \textsf{APAMT}~\cite{liu2020learning}: Attentional Profile-Aware Multi-Task model (APAMT) is proposed in our preliminary work. It leverages the Profile-aware LSTM and the soft-attention mechanism to model the daily behavior sequence. Besides, it adopts a simpler multi-task learning framework compared with DAPAMT. Specifically, it implicitly models the interactions among tasks by setting shared layers and task-specific layers.
\end{itemize}

\begin{table}[b]
\caption{Dataset statistics.}
\label{tab:stat}
\centering
\begin{tabular}{l|r}
\hline
Item                                         & Value                                                                                           \\ \hline
\# Students                                  & 10,000                                                                                          \\ \cline{2-2} 
\multirow{2}{*}{Student Behaviors Time Span} & \begin{tabular}[c]{@{}r@{}}09/12/2016 - 01/15/2017 \\ (i.e., Fall 2016 semester);\end{tabular}  \\
                                             & \begin{tabular}[c]{@{}r@{}}02/20/2017 - 06/25/2017 \\ (i.e., Spring 2017 semester)\end{tabular} \\ \cline{2-2} 
\# Library Entrance Records                  & 867,571                                                                                         \\
\# Dormitory Entrance Records                & 1,783,595                                                                                       \\
\# Demographic Records                       & 10,000                                                                                          \\
\# Courses                                   & 2,482                                                                                           \\ \hline
\end{tabular}
\end{table}

\subsection{Evaluation Metric}
We adopt Mean Square Error (MSE) to quantify the distance between the predicted scores and the actual ones. MSE penalizes large errors more heavily than the non-quadratic metrics, and thus takes higher numerical values. Given our tasks, if a method performs very well for half the students and poorly for the other half, we still think it is not a good method. Hence we select MSE as the metric to compare methods. The smaller the values are, the better results the method has. The metric is defined as:
\begin{equation}
    MSE=\frac{1}{|U|}\sum_{i=1}^U(z^i-\tilde{z}^i)^2,
\end{equation}
where $i$ denotes one testing instance, $U$ is the number of testing samples, $z^i$ is the actual value, and $\tilde{z}^i$ is the predicted value. For better understanding the metric and actual data, we analyze our dataset. The minimum WAG is $26.83$ and the maximum WAG is $97.82$ in our dataset. Thus the MSE range could be between $0$ and $5039.58$ on PAP task. The minimum number of borrowed books is $0$ and the maximum number is $137$ in our dataset. Thus the MSE range could be between $0$ and $18769$ on PNBB task. The minimum number of failed courses is $0$ and the maximum number is $10$ in our dataset. Thus the MSE range could be between $0$ and $100$ on PNFC task. We can know one method is more likely to get a small MSE on PNFC task and get a big MSE on PNBB task.

\subsection{Implementation Details}
For baselines except \textsf{APAMT}, we extract features from heterogeneous student behaviors as Guan et al.~\cite{guan2015discovery} suggested. In addition, due to the uncertainty of the historical observation sequence, we extract some descriptive statistics (minimum, maximum and mean) as features. As for the data preprocessing, we represent the categorical features with one-hot encoding. We process numerical inputs with the Min-Max normalization to ensure they are within a suitable range. For WAG, the number of borrowed books and the number of failed courses, because we use $\tanh$ as the activation function in the output layer, we scale them into $[-1, 1]$. In the evaluation, we re-scale the predicted values back to the normal values, compared with the actual values. For the other numerical inputs, we scale them into $[0, 1]$. Around half of the data (Fall 2016) are used as the training set and the other half (Spring 2017) are used for testing.

The hyper-parameters of all models are tuned with a ten-fold cross-validation method on the training dataset.
We only present the optimal settings of our model are as follows. The dense embedding layer has $30$ neurons. The dimensions of the hidden states in the Profile-aware LSTMs which handle entering the library and going back to the dormitory behaviors are set as $12$ and $4$ respectively. The dimension of the hidden state in the dynamic LSTM which handles the historical WAG/number of borrowed books/number of failed courses sequence is set as $5$. The FC layer of the Multi-task Interaction Unit has $100$ neurons. The activation function used in the FC layer is $\mathrm{PReLU}$. We adopt $4$ Multi-task Interaction Units in total. We apply dropout before the output layer with a dropout rate equal to $0.4$.

\subsection{Significance Test}
The two-tailed unpaired $t$-test is performed to detect significant differences between \textsf{DAPAMT} and the best baseline. There are two conditions that should be met: the two groups of samples should be normally distributed; the variances of the two groups are the same (this can be checked using Levene's test).

\subsection{Results and Discussion}
To comprehensively evaluate our model, we conduct five experiments. First, we compare our proposed model with state-of-the-art methods to show the effectiveness of our model. Second, we do ablation studies to prove the effectiveness of the key components in \textsf{DAPAMT} such as the Profile-aware LSTM. Third, we provide the visualization of learned representations to show whether our model can learn general representations which could be transferred to some new tasks. Fourth, we randomly select some students to visualize soft-attention weights and co-attention weights. We do this experiment to see whether our model can provide interpretability and whether attention mechanisms are effective. Finally, we study how hyper-parameters in \textsf{DAPAMT} impact the performance.

\begin{table}[t]
    \caption{Comparison of different methods. The results with the best performance are marked in bold and the second-best results are shown as underlined. The column RI denotes the average relative improvement for MSE on DAPAMT over Baselines. $^\star$ represents significance level $p$-value $<0.05$ of comparing DAPAMT with the best baseline.}
    \label{tab:re}
    \centering
    \begin{tabular}{l|ll|ll|ll}
    \hline
    \multirow{2}{*}{Compared Methods} & \multicolumn{2}{c|}{PAP} & \multicolumn{2}{c|}{PNBB} & \multicolumn{2}{c}{PNFC} \\ \cline{2-7} 
                                      & MSE            & RI      & MSE             & RI      & MSE            & RI      \\ \hline
    \textsf{HA}                                & 31.85          & 61.82\% & 63.50           & 65.95\% & 0.344          & 54.94\% \\
    \textsf{LSTM}                              & 28.46          & 57.27\% & 57.21           & 62.21\% & 0.319          & 51.41\% \\
    \textsf{BRR}                               & 27.68          & 56.07\% & 50.06           & 56.81\% & 0.306          & 49.35\% \\
    \textsf{SVR}                               & 26.81          & 54.64\% & 49.83           & 56.61\% & 0.275          & 43.64\% \\
    \textsf{RF}                                & 18.24          & 33.33\% & 30.17           & 28.34\% & 0.197          & 21.32\% \\
    \textsf{GBDT}                              & 18.87          & 35.56\% & 29.73           & 27.28\% & 0.220          & 29.55\% \\
    \textsf{MLP}                               & 17.72          & 31.38\% & 28.64           & 24.51\% & 0.183          & 15.30\% \\
    \textsf{APAMT}                             & \underline{14.52}    & 16.25\% & \underline{24.77}     & 12.72\% & \underline{0.167}    & 7.19\%  \\ \hline
    \textsf{DAPAMT}                            & $\textbf{12.16}^\star$ & -       & $\textbf{21.62}^\star$  & -       & $\textbf{0.155}^\star$ & -       \\ \hline
    \end{tabular}
\end{table}

\subsubsection{Exp-1: Comparison with Baselines.}
\ 
\newline
We compare our model with baselines. Quantitative comparison between different models on the three prediction tasks is shown in Table~\ref{tab:re}. All baselines except \textsf{APAMT} are trained with single task. From the table, we can see that \textsf{DAPAMT} achieves the best performances with the lowest MSE $12.16$ on PAP task, the lowest MSE $21.62$ on PNBB task and the lowest MSE $0.155$ on PNFC task.

We can see that \textsf{HA} performs worst since the trend of the WAG/number of borrowed books/number of failed courses change is not so stable. This indicates that it is necessary to develop a method to predict the future trend. We can see that \textsf{LSTM} performs not well, as it only utilizes the historical WAG/number of borrowed books/number of failed courses values. Other baselines further consider more information, such as student profile information and heterogeneous student behaviors, and therefore achieve better performances. Although \textsf{BRR} performs better than \textsf{LSTM}, it still shows poor performance which indicates that capturing linear correlations is not sufficient. For \textsf{SVR}, we leverage RBF kernel thus it can model the non-linear relation between features. As a result, \textsf{SVR} performs better than \textsf{BRR}. Ensemble models (i.e., \textsf{RF} and \textsf{GBDT}) perform well. Ensemble methods usually outperform single models so they are popular. \textsf{MLP} performs best among baselines except for \textsf{APAMT}. Since it can model complex relations among features with the help of deep learning. \textsf{APAMT} performs best among baselines because it can not only extract features automatically from heterogeneous behaviors but also model the interactions among multiple tasks in an implicit way. 

It is worth mentioning that \textsf{DAPAMT} achieves $16.2\%$, $12.7\%$, $7.2\%$ relative improvements on PAP, PNBB, PNFC tasks, compared with the best baseline, ie., \textsf{APAMT}. One important reason is that \textsf{DAPAMT} can explicitly model the interactions among multiple tasks. With the help of the co-attention mechanism, \textsf{DAPAMT} can control the knowledge transfer among multiple tasks. 

\begin{table}[t]
    \caption{Comparison with Variants of DAPAMT.}
    \label{tab:as}
    \centering
    \begin{tabular}{l|c|c|c}
    \hline
    \multirow{2}{*}{Methods}                        & PAP            & PNBB           & PNFC           \\
                                                    & MSE            & MSE            & MSE            \\ \hline
    \textsf{APAMT Trained w. Single Task}           & 15.10          & 26.43          & 0.177          \\
    \textsf{DAPAMT w. Standard LSTM}                & 13.13          & 24.42          & 0.161          \\
    \textsf{DAPAMT w/o Soft-attention Mechanism}    & 13.15          & 24.89          & 0.167          \\
    \textsf{APAMT}                                  & 14.52          & 24.77          & 0.167          \\
    \textsf{DAPAMT}                                 & 12.16          & 21.62          & 0.155          \\ \hline
    \end{tabular}
\end{table}

\subsubsection{Exp-2: Ablation Studies.} \label{as-sec}
\ 
\newline
Table~\ref{tab:as} provides the comparison results of variants of our proposed method. 

First, we study the effectiveness of our proposed multi-task learning framework. We train \textsf{DAPAMT} with single task. In this way, \textsf{DAPAMT} becomes \textsf{APAMT Trained w. Single Task}. From the table we can see that \textsf{DAPAMT} achieves $19.5\%$, $18.2\%$, $12.4\%$ relative improvements on PAP, PNBB, PNFC tasks. It demonstrates the benefits of the multi-task learning framework. Multi-task learning allows transferring of useful knowledge among related tasks. Second, we prove the benefit of the Profile-aware LSTM. We replace the Profile-aware LSTM with the standard LSTM. The result is that \textsf{DAPAMT} achieves lower MSE values (a reduction of $7.4\%$, $11.5\%$, and $3.7\%$, respectively) with the help of the Profile-aware LSTM. Profile-aware LSTM can capture profile information when modeling the daily behavior sequence. Third, we remove the soft-attention mechanism and the performances become much poorer. The improvements of \textsf{DAPAMT} are about $7.5\%$, $13.1\%$, $7.2\%$ on PAP, PNBB, PNFC tasks. This demonstrates the effectiveness of our soft-attention mechanism. Our soft-attention mechanism can dynamically learn the different importance degrees of different days for every student. Finally, compared with \textsf{APAMT}, we can conclude that our co-attention mechanism is effective. Our co-attention mechanism can explicitly control the knowledge transfer.

\begin{figure}[t]
\centering
    \subfigure[]{
        \includegraphics[width= 6.8cm]{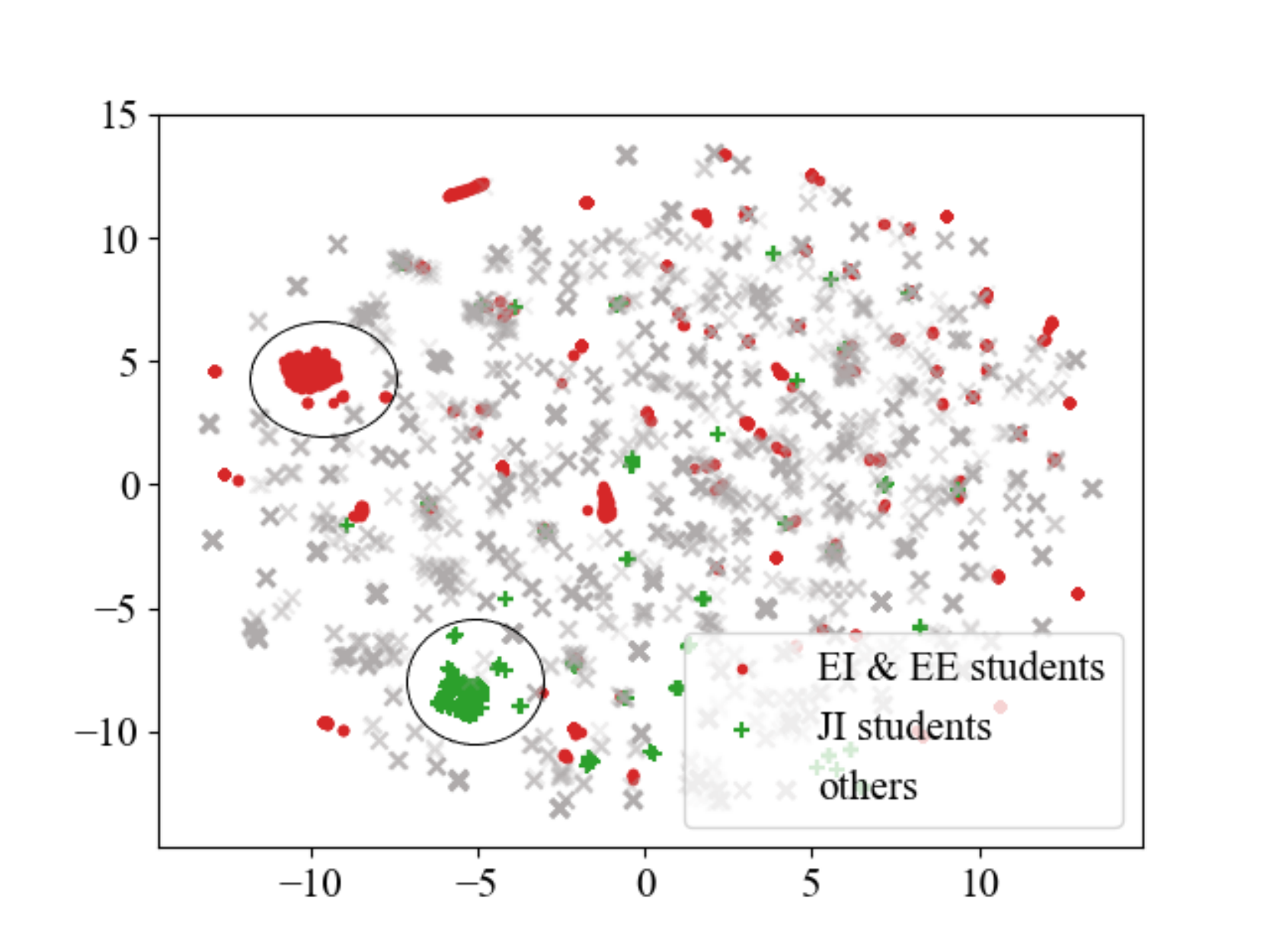}}
    \subfigure[]{
        \includegraphics[width= 6.8cm]{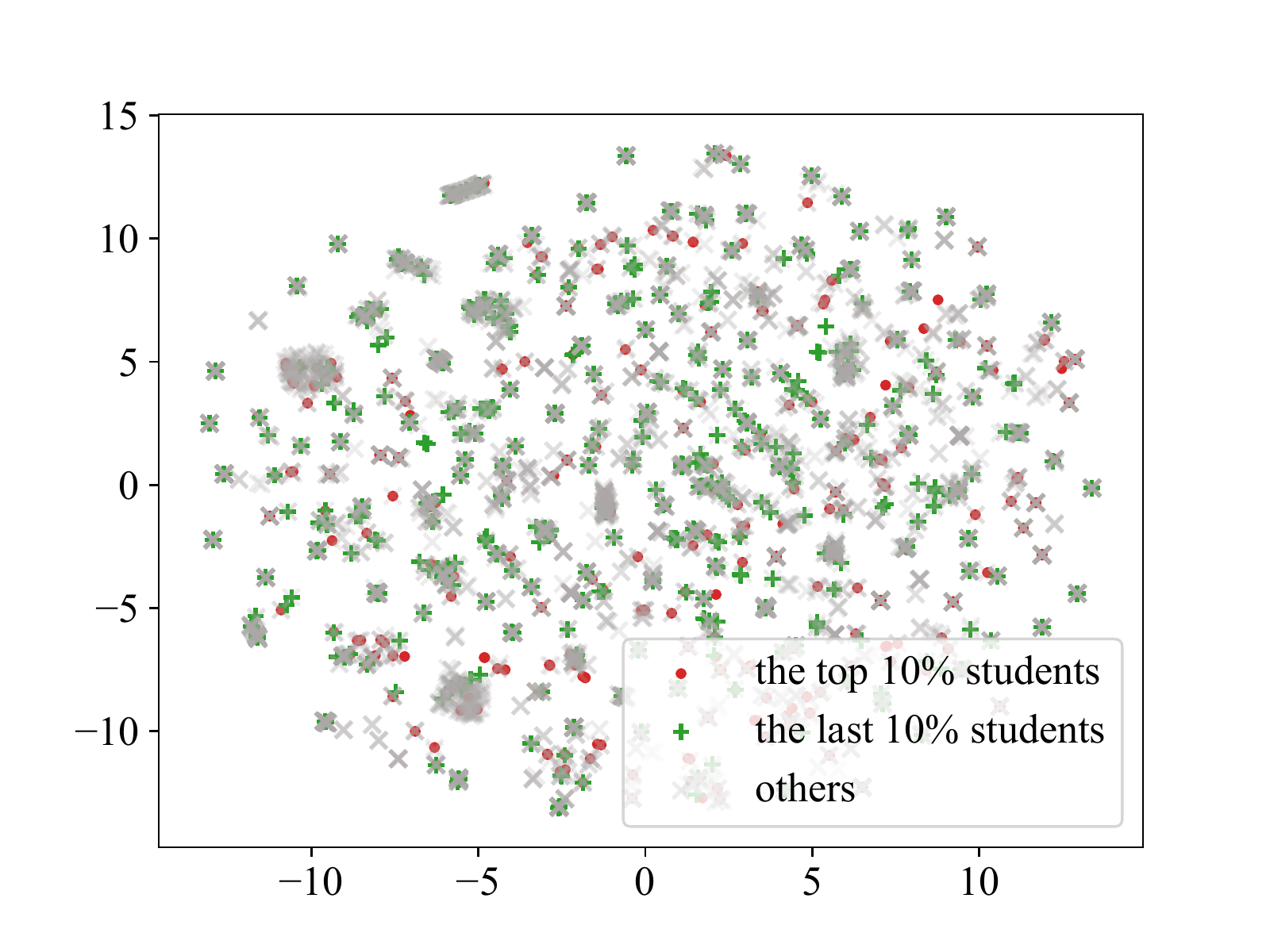}}
    \subfigure[]{
        \includegraphics[width= 6.8cm]{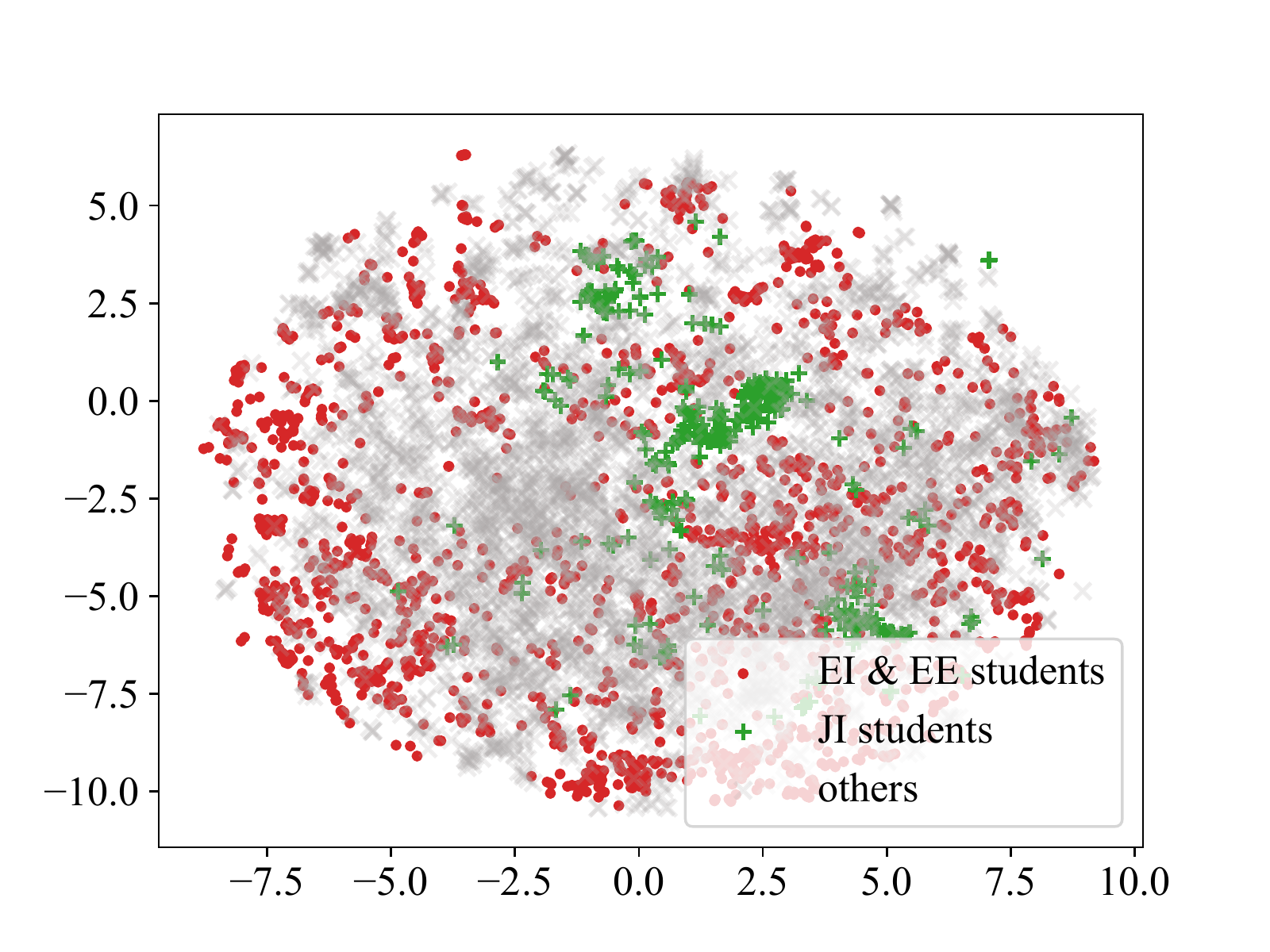}}
    \subfigure[]{
        \includegraphics[width= 6.8cm]{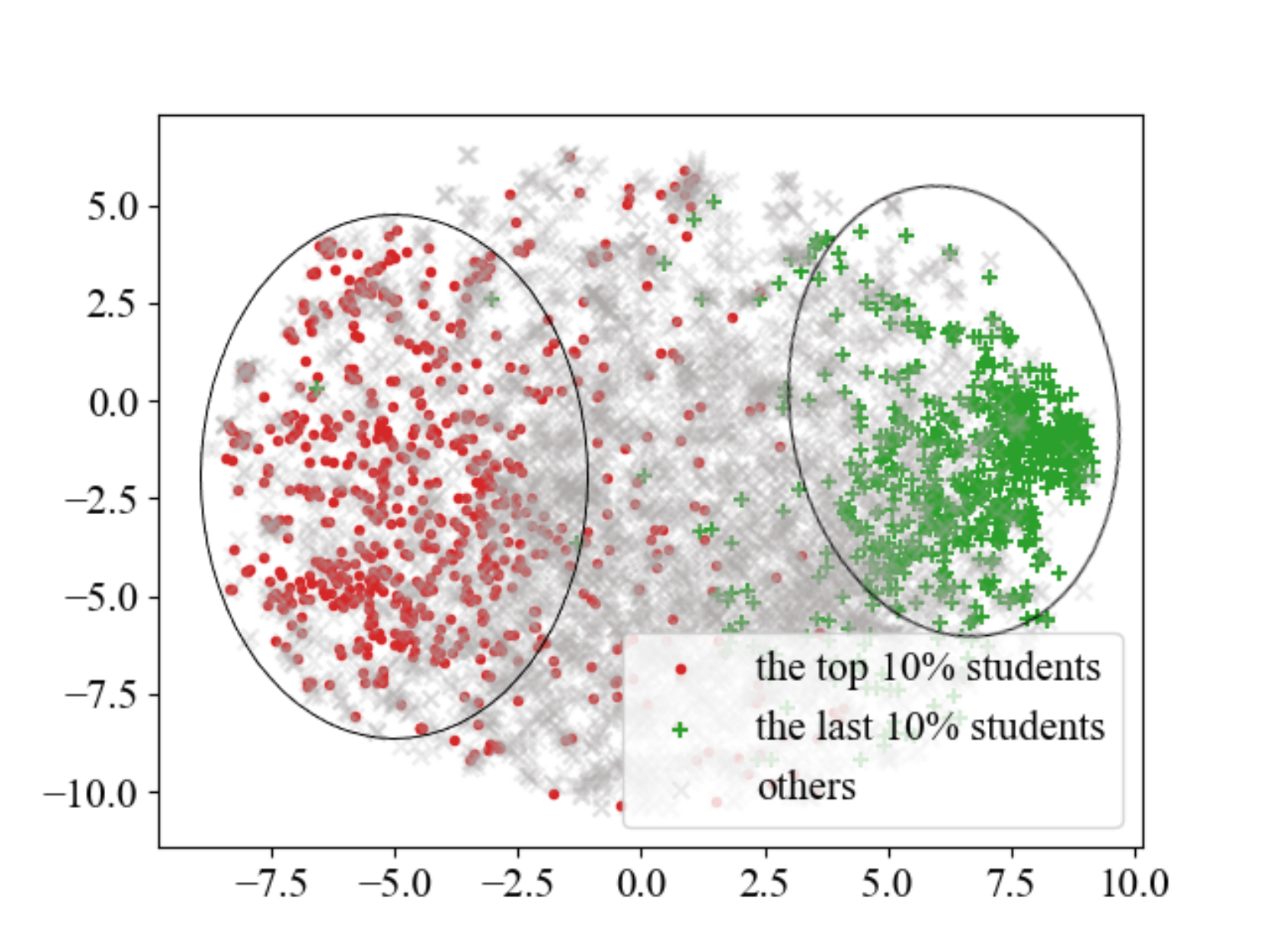}}
\caption{(a) and (b) show visualizations of learned student representations $\bm{R}$. (c) and (d) show visualizations of learned PAP task-specific representations after the first two Multi-task Interaction Units.}
\label{fig:vr}
\end{figure}

\subsubsection{Exp-3: Learned Representations Visualization.} 
\ 
\newline
We remove course information (i.e., $\bm{\bar{e}}$ and $\bm{\bar{v}}$) from \textsf{DAPAMT} and retrain \textsf{DAPAMT} with the three tasks. Figure~\ref{fig:vr} (a) and (b) show visualizations of learned student representations $\bm{R}$ based on the testing set. Figure~\ref{fig:vr} (c) and (d) show visualizations of learned PAP task-specific representations after the first two Multi-task Interaction Units based on the testing set. The technique we use is t-SNE algorithm~\cite{maaten2008visualizing}. Specifically, with learned representations as input, t-SNE algorithm maps students to the 2-D space. One dot in the figure represents one student.

We randomly choose two groups of students (EI \& EE (i.e., Electronic Information and Electrical Engineering) students and JI (i.e., Joint Institute) students) and mark these students. From Figure~\ref{fig:vr} (a), we can clearly find two clusters of EI \& EE students and JI students. But we can not clearly find two clusters in Figure~\ref{fig:vr} (c). We can only distinguish these two groups.

\begin{figure*}[t]
\centering
    \includegraphics[width= 11cm]{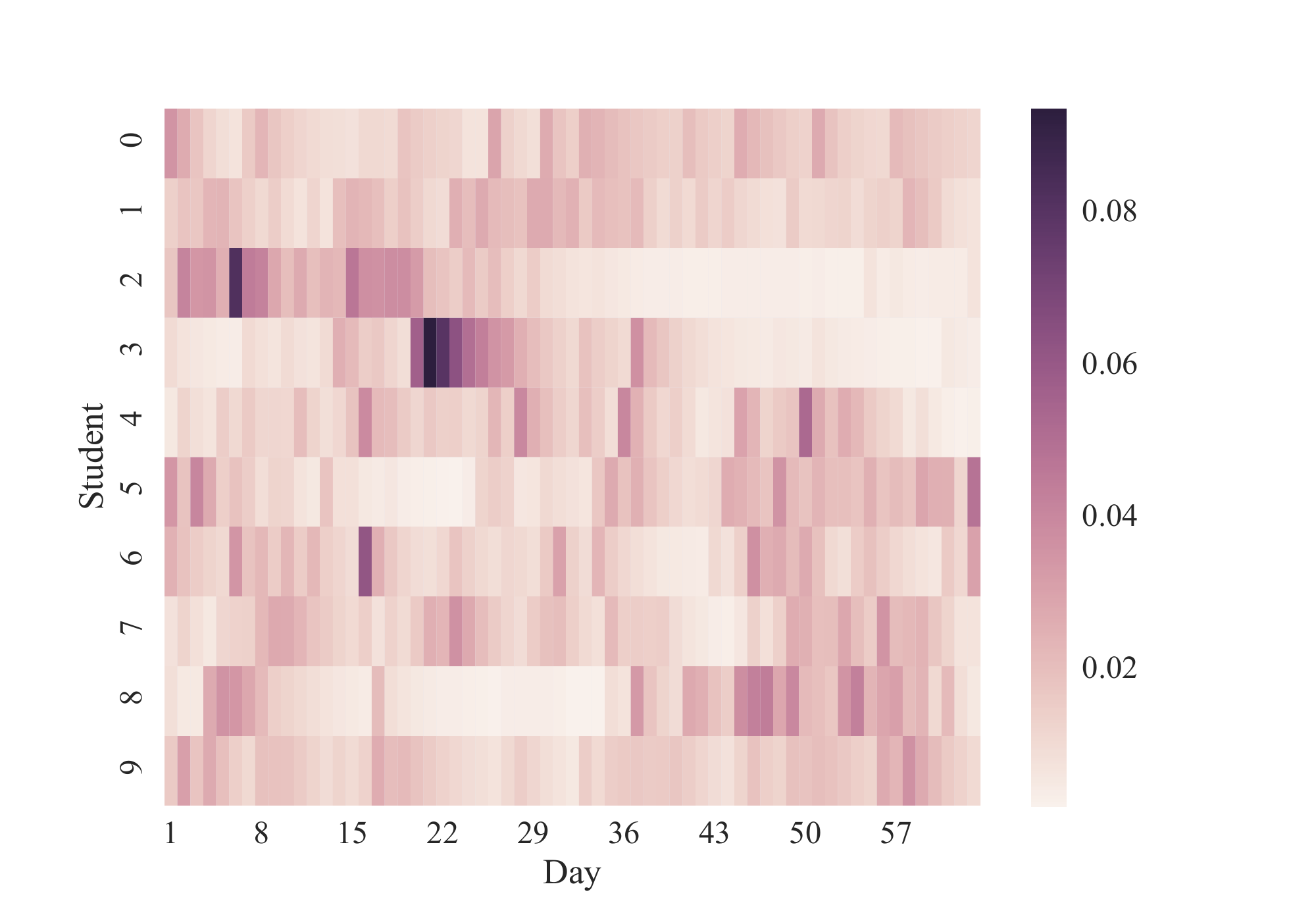}
    \caption{Visualization of soft-attention weights.}
    \label{fig:vsatt}
\end{figure*}
\begin{figure*}[t]
\centering
    \includegraphics[width= 11cm]{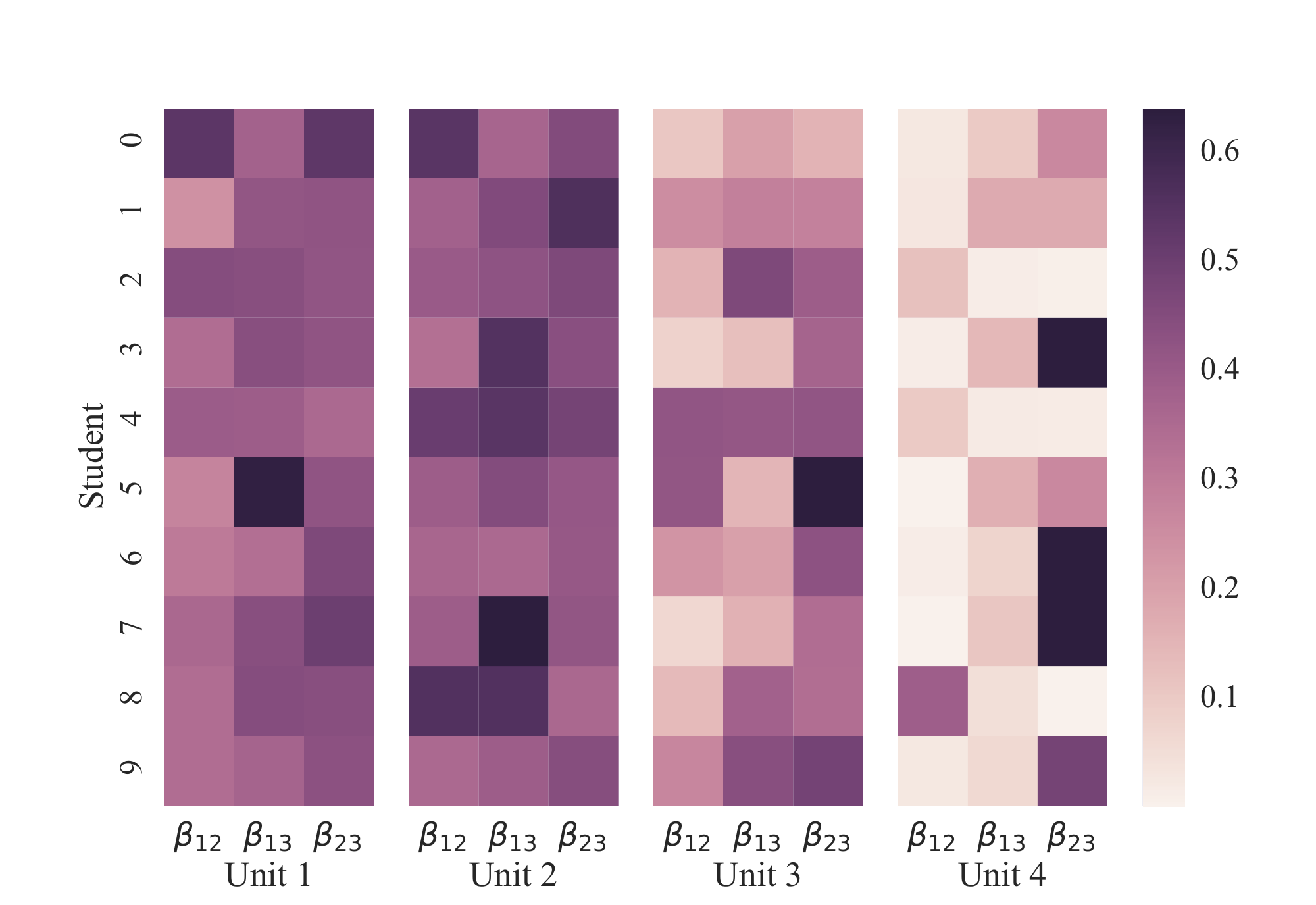}
    \caption{Visualization of co-attention weights.}
    \label{fig:vcatt}
\end{figure*}

Next, we mark the top $10\%$ students and the last $10\%$ students according to the mean of all WAGs. From Figure~\ref{fig:vr} (b), we can not distinguish these two groups of students. But there are clearly two clusters in Figure~\ref{fig:vr} (d).

This experiment indicates student representation $\bm{R}$ is general and could be transferred to other tasks. Besides, after seeing the visualization of learned task-specific representations, we can know our model performs well and generates a meaningful layout of students (students with similar WAGs are distributed closer).

\begin{figure}[t]
\centering
    \subfigure[]{
        \includegraphics[width= 11.5cm]{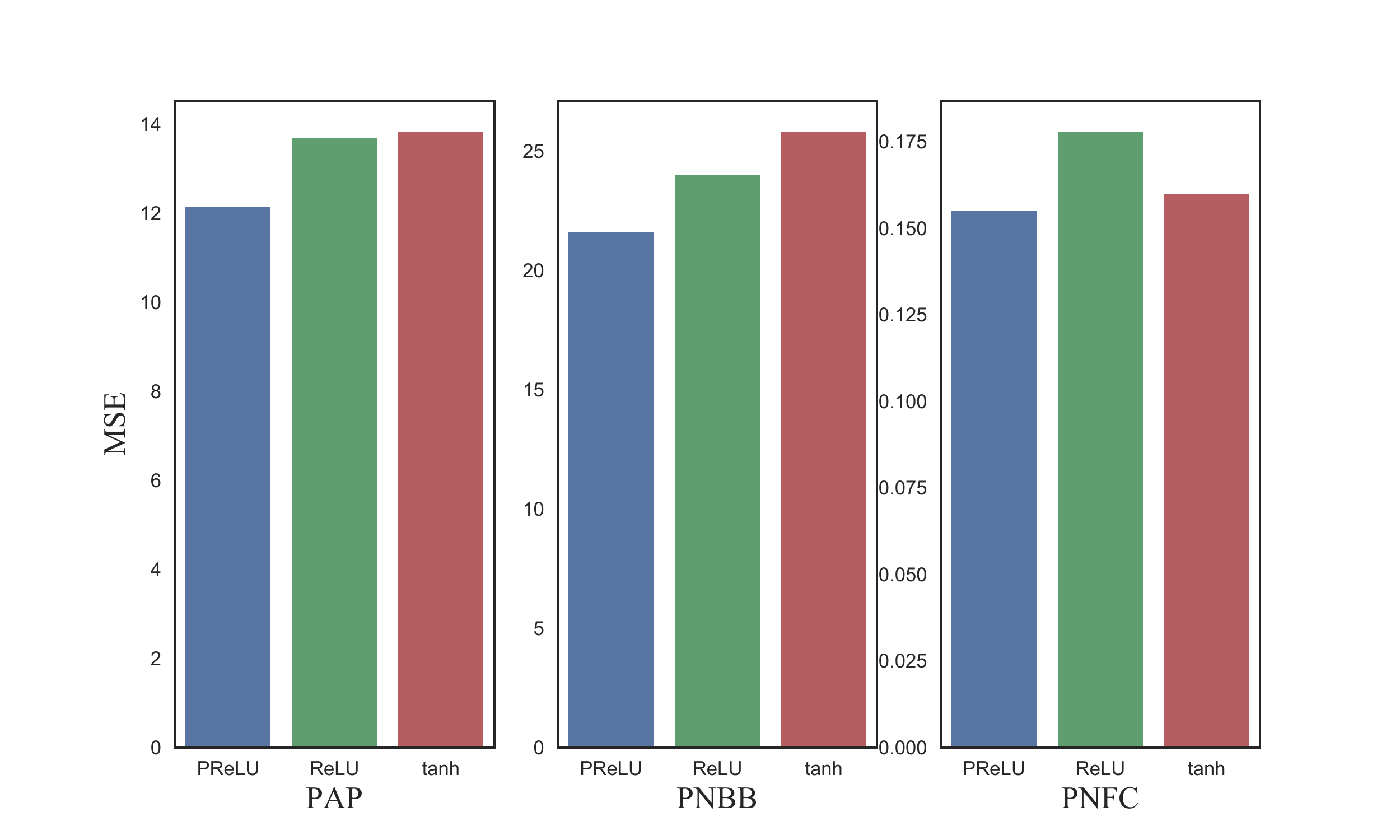}}
    \subfigure[]{
        \includegraphics[width= 11.5cm]{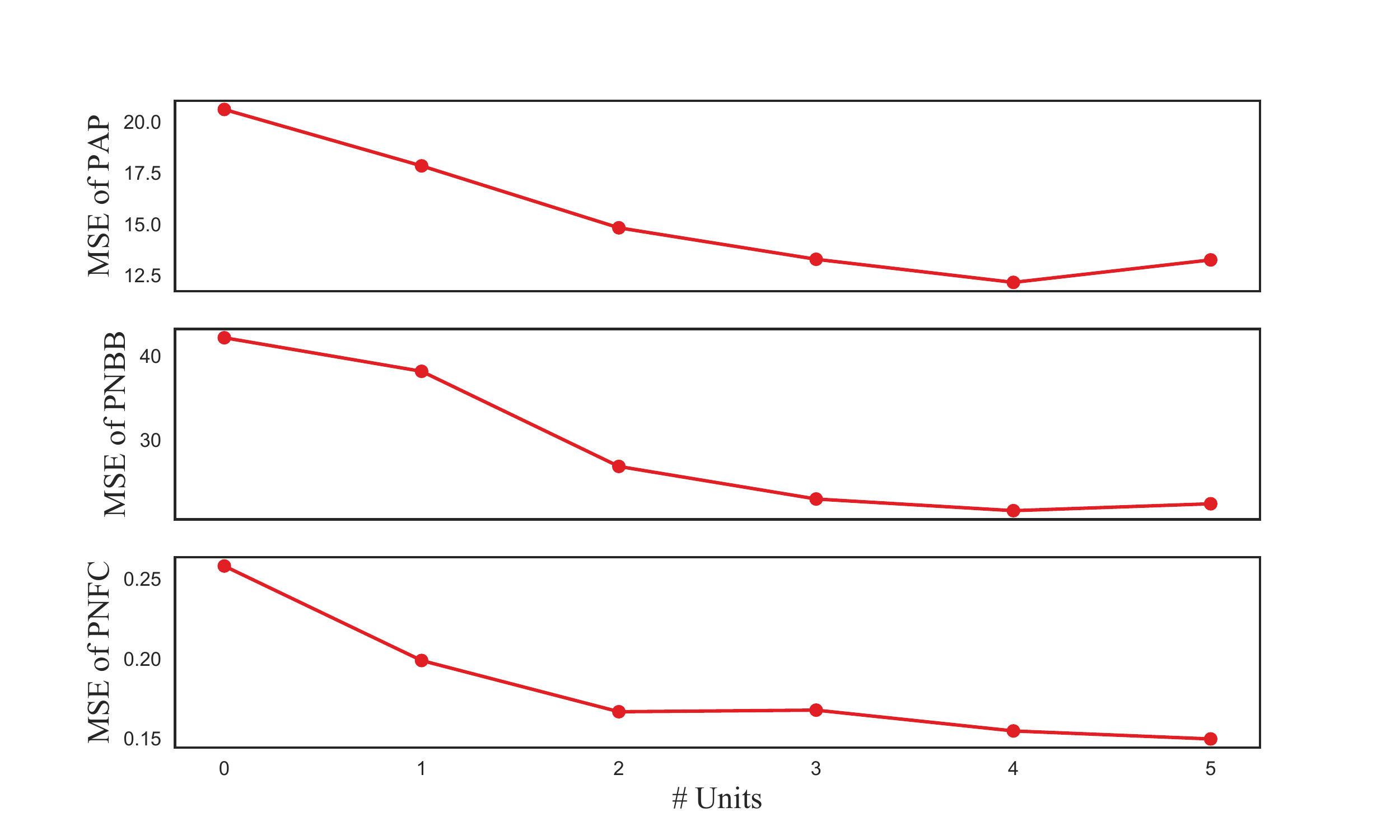}}
\caption{MSE with activation functions (a) and \# units (b).}
\label{fig:hp}
\end{figure}

\subsubsection{Exp-4: Attention mechanisms Visualization.}
\ 
\newline
\noindent\underline{\emph{Exp-4.1: Soft-attention Visualization.}} We randomly select $10$ students in the testing set and visualize the soft-attention weights of days in Figure~\ref{fig:vsatt} to show whether \textsf{DAPAMT} can find out informative days for different students. The figure shows that soft-attention weights vary for days and students. 

Next, we choose student $2$ to do the case study. For simplicity, we only show student $2$'s library entrance records. The fragment of the records is \{(stu2,2017-02-22 15:21:54),(stu2,2017-02-25 15:58:29),(stu2,2017-04-15 12:07:32),(stu2,2017-04-16 08:52:53)\}. In other words, during the first $63$ days, student $2$ enters the library on day $3$, $6$, $55$, and $56$. From the figure, we can see that the weights of these days are large. 

This experiment visually indicates our designed soft-attention mechanism is effective (this conclusion can also be got in Section~\ref{as-sec}) and gives the informative day a large weight.

\noindent\underline{\emph{Exp-4.2: Co-attention Visualization.}} We still use the $10$ students and visualize the co-attention weights. In particular, we visualize the co-attention weights between PAP task and PNBB task (i.e., $\beta_{12}$), the weights between PAP task and PNFC task (i.e., $\beta_{13}$), and the weights between PNBB task and PNFC task (i.e., $\beta_{23}$) with the number of stacked units varying from $1$ to $4$. We choose student $3$ and student $8$ to do the case studies. The WAG of student $3$ is $87$, but the number of books borrowed by student $3$ is $0$. So the weights $\beta_{12}$ of student $3$ are pretty small. The number of courses in which student $3$ failed is $0$. Thus the weights $\beta_{23}$ of student $3$ are large. Next, we focus on student $8$. The WAG of student $8$ is $82.33$, and the number of books borrowed by student $8$ is $33$. So the weights $\beta_{12}$ of student $8$ are large. The number of courses in which student $8$ failed is $0$. Thus the weights $\beta_{23}$ of student $8$ become very small in unit 4.

This experiment visually indicates our designed co-attention mechanism is effective (this conclusion can also be got in Section~\ref{as-sec}) and explicitly controls the knowledge transfer among tasks by weights.


\subsubsection{Exp-5: Effect of Hyper-parameters.}
\ 
\newline
\noindent\underline{\emph{Exp-5.1: Effect of Activation Functions.}} We investigate the influence of the activation function which exists in FC layers. We choose PReLU, ReLU, and tanh to do the experiment. Figure~\ref{fig:hp} (a) shows the results. We observe that PReLU is more suitable.

\noindent\underline{\emph{Exp-5.2: Effect of the Number of Multi-task Interaction Units.}} A problem worth studying is that how many units are appropriate. Experimental results are shown in Figure~\ref{fig:hp} (b). As the number of units grows, the performances grow. When the number of units is $4$, the performances except on PNFC drop. We think the reason might lie in the gradient vanishing or the overfitting problem as the whole network goes deeper. So we adopt $4$ stacked units.

\section{Ethical Considerations}\label{ec}
The ethical implication is our major concern. Nicholson and Glenn~\cite{price2019privacy} proposed that the ethical analysis mainly depends on three aspects: the types of the data; the people who will be accessing the data; and the purpose of the work. So in what follows, we do the ethical analysis in three aspects.

For the types of data, the dataset used in our work has been processed and fully anonymized. Student names are removed. Sensitive information (i.e., student ID, place of birth, nationality, gender) is encrypted by being hashed to a vector space.

For the people who will be accessing the data, the data used in our work are not released for public access. In applications of our model, we stress that only college official management offices collect required data and prediction results should be constrained in a small group of official college staff.

For the purpose of this work, we would like to stress that the use of student data and prediction results should be constrained for the purpose of college student management only, not for general public access. Besides, the prediction results should not bias an instructor's treatment of individual students. In the real world, machine learning algorithms will never achieve 100\% accurate prediction results~\cite{khan2020student}.

In addition, this work has been approved by the Institutional Review Board (IRB).

\section{Conclusion}\label{co}
In this paper, we propose a Dual Attention Profile-Aware Multi-Task model (i.e., DAPAMT) to jointly modeling heterogeneous student behaviors generated from digital footprints and interactions among multiple prediction tasks. With DAPAMT, we can learn personalized and general student representations from student profiles and student heterogeneous behaviors. At the same time, we can explicitly control the knowledge transfer among prediction tasks. Qualitative and quantitative experiments on a real-world dataset have demonstrated the effectiveness of DAPAMT. We believe DAPAMT is an extensible framework. DAPAMT can be utilized to model heterogeneous behaviors of one person rather than one student and can be utilized to handle more tasks in actual scenarios.

\section*{Acknowledgements}
We thank the anonymous reviewers for carefully reviewing and useful suggestions. This research is supported in part by the 2030 National Key AI Program of China 2018AAA0100503 (2018AAA0100500), National Science Foundation of China (No. 62072304, No. 61772341, No. 61472254, No. 61770238), Shanghai Municipal Science and Technology Commission (No. 18511103002, No. 19510760500, and No. 19511101500), the Program for Changjiang Young Scholars in University of China, the Program for China Top Young Talents, the Program for Shanghai Top Young Talents, SJTU Global Strategic Partnership Fund (2019 SJTU-HKUST), the Oceanic Interdisciplinary Program of Shanghai Jiao Tong University (No. SL2020MS032) and Scientific Research Fund of Second Institute of Oceanography (No. SL2020MS032).


\bibliographystyle{ACM-Reference-Format}


\end{document}